\documentclass[10pt,twocolumn,letterpaper]{article}

\usepackage[pagenumbers]{iccv} 
\usepackage{array}      
\usepackage{tabularx}   
\usepackage{xcolor}     
\usepackage{colortbl}   
\usepackage{booktabs}   
\usepackage{multirow}
\usepackage{subcaption}

\definecolor{gray}{rgb}{0.95,0.95,0.95}
\definecolor{iccvblue}{rgb}{0.21,0.49,0.74}
\usepackage[pagebackref,breaklinks,colorlinks,allcolors=iccvblue]{hyperref}

\newcommand{\mymodel}{CODiff}

\def\paperID{824} 
\def\confName{ICCV}
\def\confYear{2025}

\title{Compression-Aware One-Step Diffusion Model for JPEG Artifact Removal}

\author{Jinpei Guo$^{1,2}$,\; Zheng Chen$^{2}$,\; Wenbo Li$^{3}$,\; Yong Guo$^{4}$,\; Yulun Zhang$^{2}$\thanks{Corresponding author: Yulun Zhang, yulun100@gmail.com}\\
$^{1}$Carnegie Mellon University,\; $^{2}$Shanghai Jiao Tong University, \\$^{3}$The Chinese University of Hong Kong,\; $^{4}$Max Planck Institute for Informatics}

\begin{document}
\maketitle

\begin{abstract}
Diffusion models have demonstrated remarkable success in image restoration tasks. However, their multi-step denoising process introduces significant computational overhead, limiting their practical deployment. Furthermore, existing methods struggle to effectively remove severe JPEG artifact, especially in highly compressed images. To address these challenges, we propose \mymodel, a \textbf{c}ompression-aware \textbf{o}ne-step \textbf{diff}usion model for JPEG artifact removal. The core of \mymodel~is the compression-aware visual embedder (CaVE), which extracts and leverages JPEG compression priors to guide the diffusion model. Moreover, We propose a dual learning strategy for CaVE, which combines explicit and implicit learning. Specifically, explicit learning enforces a quality prediction objective to differentiate low-quality images with different compression levels. Implicit learning employs a reconstruction objective that enhances the model's generalization. This dual learning allows for a deeper understanding of JPEG compression. Experimental results demonstrate that \mymodel~surpasses recent leading methods in both quantitative and visual quality metrics. The code is available at \href{https://github.com/jp-guo/CODiff}{https://github.com/jp-guo/CODiff}.
\end{abstract}
\vspace{-8mm}

\setlength{\abovedisplayskip}{2pt}
\setlength{\belowdisplayskip}{2pt}

\section{Introduction}
\label{intro}

\begin{figure}[t]
\scriptsize
\centering
\begin{tabular}{cc}
\hspace{-0.55cm}
\begin{tabular}{c}
\includegraphics[width=0.172\textwidth, height=0.185\textwidth]{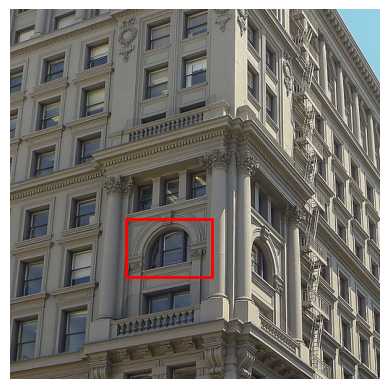}
\\
Urban100: img\_014 \\
MACs (T) / Time (s)
\end{tabular}
\hspace{-6mm}
\begin{tabular}{cccccc}
\includegraphics[width=0.105\textwidth]{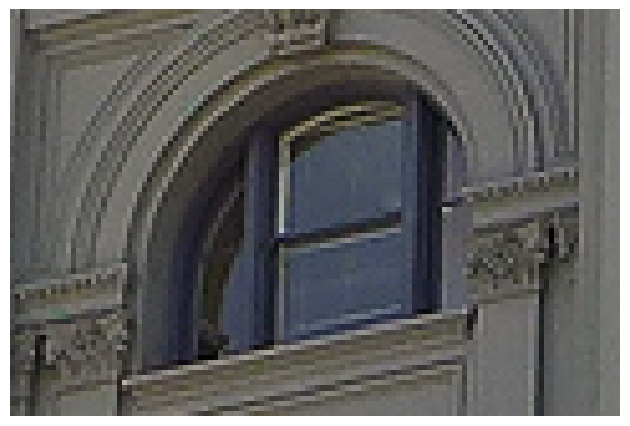} \hspace{-5mm} &
\includegraphics[width=0.105\textwidth]{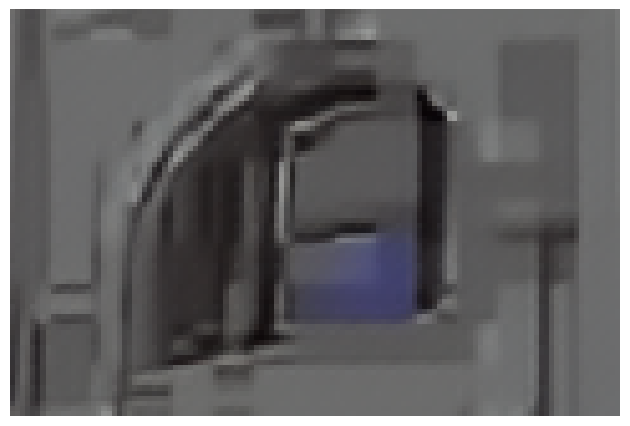} \hspace{-5mm} &
\includegraphics[width=0.105\textwidth]{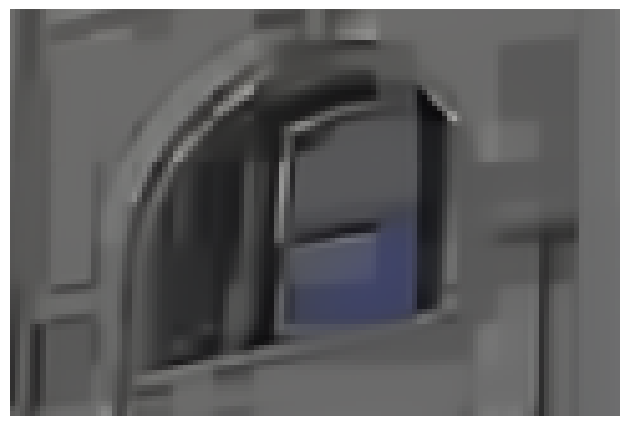} \hspace{-5mm} 
\\
HQ \hspace{-5mm}
 &
JDEC~\cite{han2024jdec} \hspace{-5mm}  &
PromptCIR~\cite{li2024promptcir} \hspace{-5mm} &
\\
 \hspace{-5mm} &
3.40 / 0.86 \hspace{-5mm}  &
30.27 / 27.97 \hspace{-5mm} &
\\
\includegraphics[width=0.105\textwidth]{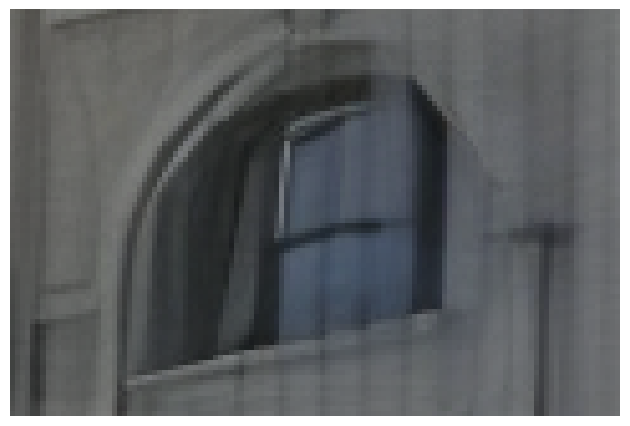} \hspace{-5mm} &
\includegraphics[width=0.105\textwidth]{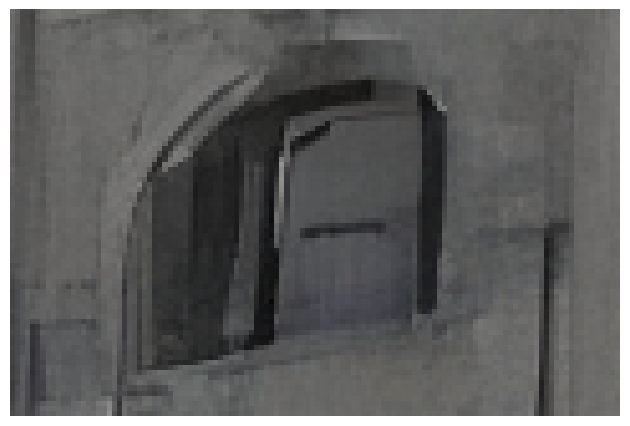} \hspace{-5mm} &
\includegraphics[width=0.105\textwidth]{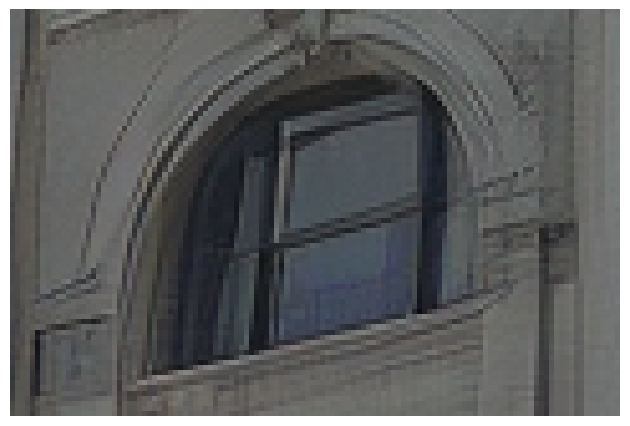} \hspace{-5mm}  
\\ 
DiffBIR*~\cite{lin2024diffbir} \hspace{-5mm}  &
OSEDiff*~\cite{wu2024one} \hspace{-5mm} &
\mymodel~(ours) \hspace{-5mm}
\\
188.24 / 50.81 \hspace{-5mm}  &
10.39 / 0.65 \hspace{-5mm} &
9.46 / 0.57 \hspace{-5mm}
\\
\end{tabular}
\vspace{-1mm}

\end{tabular}
\vspace{-2.5mm}
\caption{Visual comparison of JPEG artifact removal (QF=5). We provide multiply accumulate operations (MACs) and time during inference. DiffBIR*~\cite{lin2024diffbir}, OSEDiff*~\cite{wu2024one} are retrained on the same dataset as our method. All models are tested with an input image size of 1,024$\times$1,024. Our \mymodel~ reconstructs a more realistic and faithful visual result than competing methods.}
\label{fig:head}
\vspace{-7mm}
\end{figure}

JPEG~\cite{wallace1991jpeg} artifact removal task aims to remove the artifact caused by the compression algorithm and recover the lost information from the compressed images. Recent advancements have focused on leveraging CNN-based and Transformer-based methods~\cite{dong2015compression,zhang2019residual,liang2021swinir,jiang2021towards,li2024promptcir} to remove JPEG artifact for compressed images with different quality factors (QFs), achieving remarkable performance. However, these methods (see Fig.~\ref{fig:head}) face significant challenges at high compression levels, due to substantial compression artifact and severe visual information loss.

Recent advances in diffusion models~\cite{song2020score,ho2020denoising,song2020denoising,dhariwal2021diffusion}, particularly large-scale pre-trained text-to-image (T2I) models~\cite{rombach2022high,podell2023sdxl,saharia2022photorealistic}, have strong image generation priors. These priors make diffusion models a promising solution for compression artifact removal, especially in cases of severe degradation. However, their multi-step denoising process introduces substantial computational overhead. To address this, one-step diffusion (OSD) models~\cite{wu2024one,li2024distillation} have emerged as efficient alternatives. By leveraging large-scale pre-trained multi-step T2I diffusion models~\cite{rombach2022high,podell2023sdxl}, these OSD models strike a balance between strong restoration capabilities and significantly faster inference, making them a compelling choice for JPEG artifact removal.

As illustrated in Fig.~\ref{fig:head}, for highly compressed images (\eg, QF=5), diffusion-based models generally outperform traditional CNN-based and Transformer-based methods in reducing JPEG artifact such as blocky patterns and color banding. However, the restored images remain inharmonious, with certain artifact still visible. The primary issue lies in the insufficient integration of JPEG compression priors. Without this information, models struggle to differentiate between compression-induced distortions and natural image features, leading to suboptimal restoration quality. Notably, most existing diffusion-based methods~\cite{lin2024diffbir,yu2024scaling,wu2024seesr,wu2024one} for image restoration tasks neglect such compression-related information.

Thus, it is worthwhile to investigate how to extract JPEG compression priors and integrate them into the diffusion model. A major challenge is how to design effective JPEG compression representations. Previous works~\cite{ehrlich2020quantization,han2024jdec} use the quantization matrix~\cite{wallace1991jpeg} to represent the compression priors. However, since it only depends on the quality factor (QF) with static numerical values, the information it provides is insufficient. To dynamically utilize the JPEG priors, QF learning approaches~\cite{jiang2021towards,wang2021jpeg} attempt to learn compression representations. They train models to predict QF from low-quality (LQ) images and use intermediate visual embeddings as priors. However, as QF is merely a single integer, taking it as the sole learning objective constrains the model's ability to capture the comprehensive JPEG priors. Furthermore, the models may encounter generalization challenges when exposed to unseen compression levels. 

\begin{figure}[t]
    \begin{center}
    \begin{tabular}{cc}
        \hspace{-3mm}
        \includegraphics[width=0.49\columnwidth]{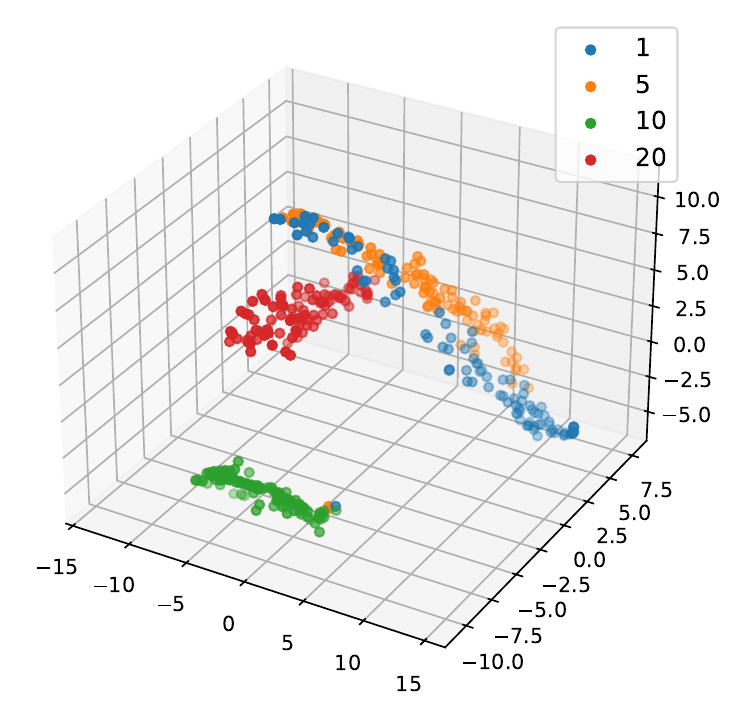}  &
        \hspace{-3mm}
        \includegraphics[width=0.49\columnwidth]{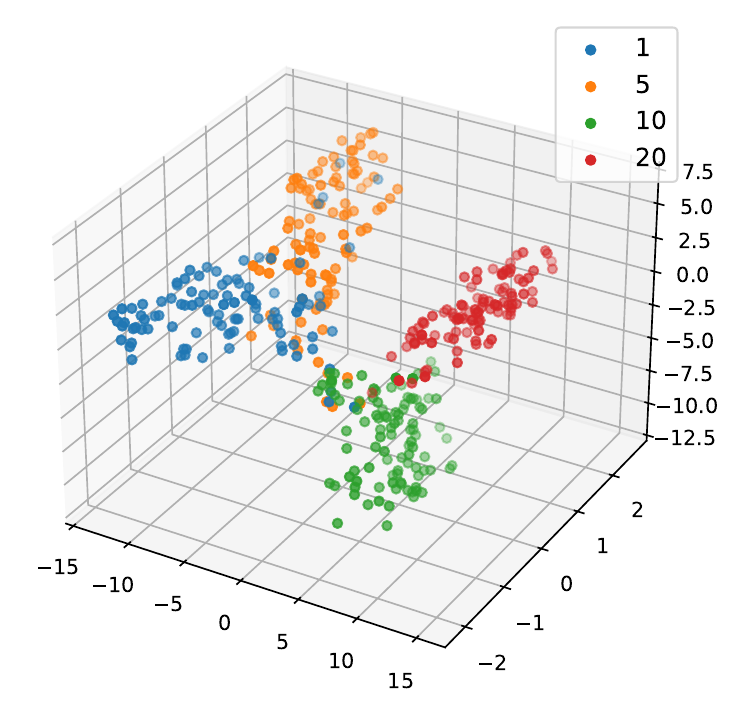}
        \vspace{-2mm}
        \\
        \hspace{-3mm}
       \scriptsize{(a) Explicit learning }& \hspace{-3mm} \scriptsize{(b) Dual learning} 
       \vspace{1mm}
       \\
       \multicolumn{2}{c}{\includegraphics[width=0.9\columnwidth]{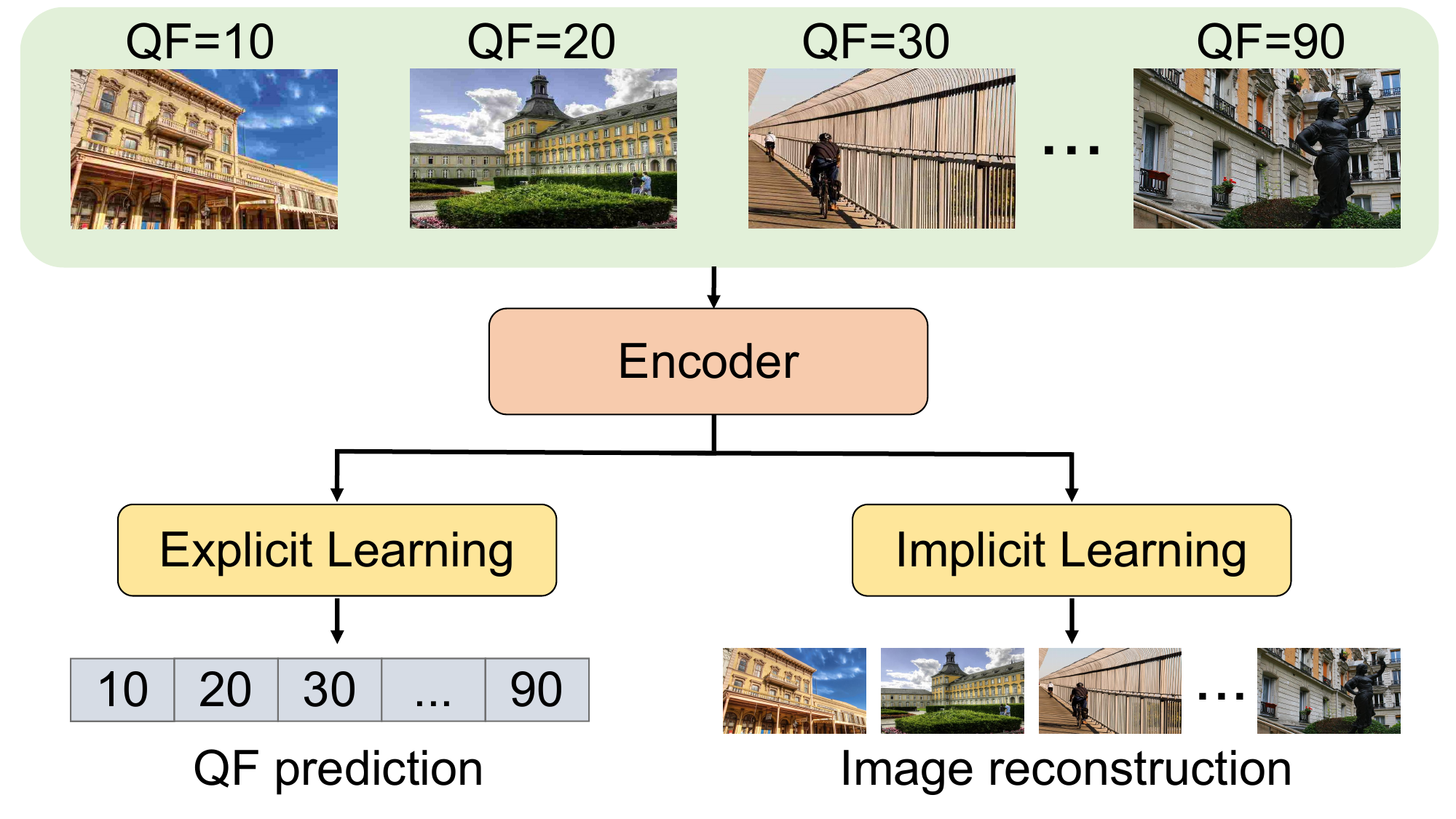}} 
       \vspace{-2mm}
       \\
       
       \multicolumn{2}{c}{\scriptsize{(c) Overview of our dual learning strategy.}}\\
       
       
    \end{tabular}
    \end{center}
    \vspace{-6mm}
    \caption{An illustration of the explicit learning approach with our dual learning strategy. We employ the t-SNE~\cite{van2008visualizing} method to visualize the clustering of output compression quality representations on the Urban100~\cite{huang2015single} dataset, where QF=1 and QF=5 are unseen compression levels. Notably, the explicit learning method struggles to distinguish between these unseen QFs, whereas our dual learning strategy effectively differentiates them.}
    \label{fig:motivation}
    \vspace{-6mm}
\end{figure}

To alleviate these challenges, we propose \mymodel, a \textbf{c}ompression-aware \textbf{o}ne-step \textbf{diff}usion model for JPEG artifact removal. \mymodel~leverages the powerful pre-trained knowledge of multi-step diffusion (MSD) models while significantly improving inference speed. A key component of \mymodel~is the compression-aware visual embedder (CaVE), which extracts JPEG compression priors to guide the denoising process. To provide more informative JPEG compression priors and improve generalization, CaVE learns the JPEG compression process through a dual-learning strategy (see Fig.~\ref{fig:motivation}). In the explicit learning phase, it is trained to predict the quality factor (QF) from a low-quality image. In the implicit learning phase, CaVE learns to restore high-quality (HQ) images from compressed inputs. This joint learning approach enhances CaVE's ability to model the JPEG compression process, thereby enhancing its generalization ability to differentiate unseen compression levels. Comparison in Fig.~\ref{fig:head} highlights the superior restoration quality and efficiency of our~\mymodel. Our main contributions are summarized as follows:
\begin{itemize}
    \item We propose \mymodel, a novel and effective OSD model for JPEG artifact removal. To the best of our knowledge, this is the first attempt to design an OSD model specifically for this task with compression priors.
    \item We design compression-aware visual embedder (CaVE), which generates embeddings that capture rich JPEG compression information. CaVE provides effective compression priors for the OSD model to restore LQ images.
    \item We propose dual learning strategy for CaVE. Specifically, explicit learning allows CaVE to distinguish LQ images across various compression levels. While implicit learning enhances CaVE's generalization capabilities.
    \item Our \mymodel~achieves SOTA performance in JPEG artifact removal, excelling in both quantitative metrics and visual quality. \mymodel~ reduces computational costs compared with both MSD and OSD models.
\end{itemize}

\vspace{-2mm}
\section{Related Work}
\vspace{-1mm}
\subsection{JPEG Artifact Removal}
\vspace{-1mm}
In recent years, learning-based methods have significantly advanced JPEG artifact removal. The pioneering work ARCNN~\cite{dong2015compression} first introduces deep learning into this task, leveraging a super-resolution network~\cite{dong2014learning} to reduce compression artifact. Transformer-based methods ~\cite{zhang2019residual,liang2021swinir} incorporate attention mechanisms to enhance feature representation. Inspired by the success of GANs in image restoration, several GAN-based methods~\cite{galteri2017deep,galteri2019deep} have been proposed to enhance the perceptual quality of compressed images. Meanwhile, dual-domain convolutional network approaches~\cite{guo2016building,kim2020agarnet,zhang2018dmcnn,zheng2019implicit,han2024jdec} have been developed to exploit redundancies in both the pixel and frequency domains. 

To further integrate the JPEG compression priors, the ranker-guided framework~\cite{wang2021jpeg} utilizes compression quality ranking as auxiliary information.
Quantization tables~\cite{li2020learning,ehrlich2020quantization} are also utilized as prior knowledge, enabling a single network to handle artifact across different quality factors (QFs). To enable blind JPEG artifact removal, FBCNN~\cite{jiang2021towards} predicts an adjustable QF to balance artifact removal and detail preservation. Additionally, PromptCIR~\cite{li2024promptcir} explores prompt learning for blind compressed image restoration. However, most existing methods struggle to recover highly compressed images due to severe information loss. With the advancement of diffusion models, incorporating JPEG compression priors into the pre-trained large scale T2I diffusion models offers a promising solution to mitigate this information loss. 

\begin{figure*}[t]
    \begin{center}
        \includegraphics[width=\textwidth]{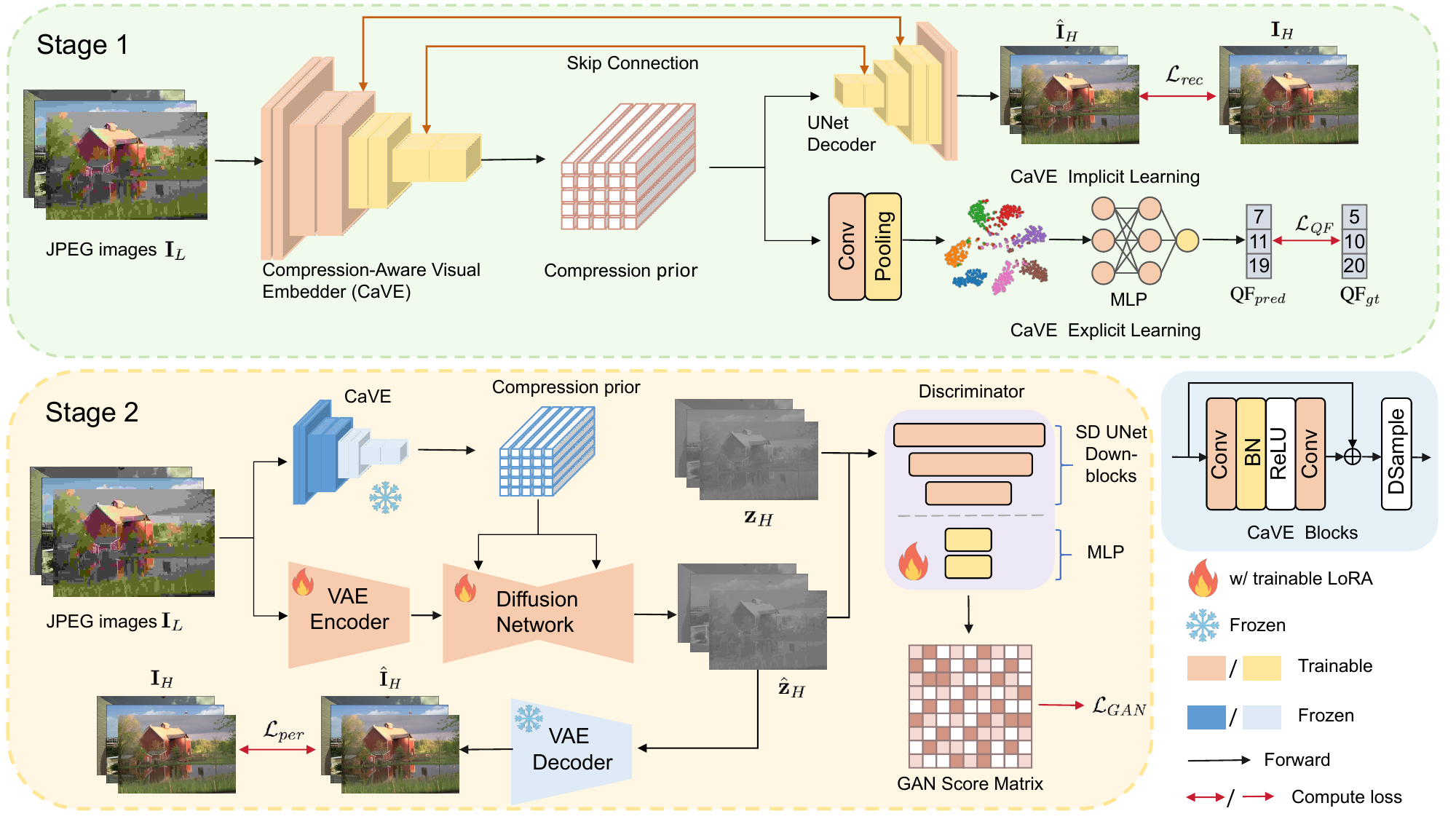}
    \end{center}
    \vspace{-20pt}
    \caption{Overview of our proposed \mymodel. In the first stage, we train our compression-aware visual embedder (CaVE) via a dual learning strategy. In implicit learning, compression prior embeddings are fed into a UNet decoder to reconstruct high-quality (HQ) images. In explicit learning, they are input into a lightweight quality factor (QF) predictor. In the second stage, the priors from CaVE are then used by the generator $\mathcal{G}_\theta$ to restore the HQ images: $\hat{\mathbf{I}}_H=\mathcal{G}_\theta(\mathbf{I}_L;\mathbf{c}_L)$. The generator $\mathcal{G}_\theta$ integrates a pre-trained VAE and UNet from StableDiffusion~\cite{rombach2022high}, with the VAE encoder and UNet fine-tuned via LoRA~\cite{hu2021lora}.} 
    \label{fig:pipeline}
    \vspace{-15pt}
\end{figure*}

\subsection{Diffusion Models}
\vspace{-2mm}
Diffusion models, known for their powerful generative capabilities, progressively transform random noise into structured data through iterative denoising. Recently, they have shown strong performance in image-to-image tasks, particularly in image restoration~\cite{wang2024exploiting,lin2024diffbir,wu2024seesr,yang2024pixel,yue2024resshift,jiang2024autodir,guo2025oscar,yu2024scaling}. By leveraging their ability to capture fine-grained details and produce high-quality outputs, diffusion models have outperformed traditional image restoration methods. However, their complex architectures~\cite{lin2024diffbir,yu2024scaling,yang2024pixel} and reliance on numerous iterative steps hinder their real-world deployment due to high computational costs.

Accelerating diffusion models by reducing inference steps is crucial for practical applications. However, excessive reduction often degrades performance. Therefore, most one-step diffusion (OSD) methods use distillation to learn from a teacher model, preserving image fidelity~\cite{yin2024dmd,yin2024dmd2,wang2024sinsr}. Notably, OSEDiff~\cite{wu2024one} has obtained promising results using variational score distillation. Despite these advances, existing diffusion-based restoration models often overlook degradation-related priors. Consequently, it is difficult to distinguish compression artifact from natural image features. Thus, exploring how to extract prior knowledge specific to JPEG compression is essential to guide diffusion models for JPEG artifact removal tasks.

\begin{figure*}[t]
    \begin{center}
    \resizebox{1\textwidth}{!}{
        \begin{tabular}{cccc}
            \hspace{-2mm}
            \includegraphics[width=0.25\textwidth]{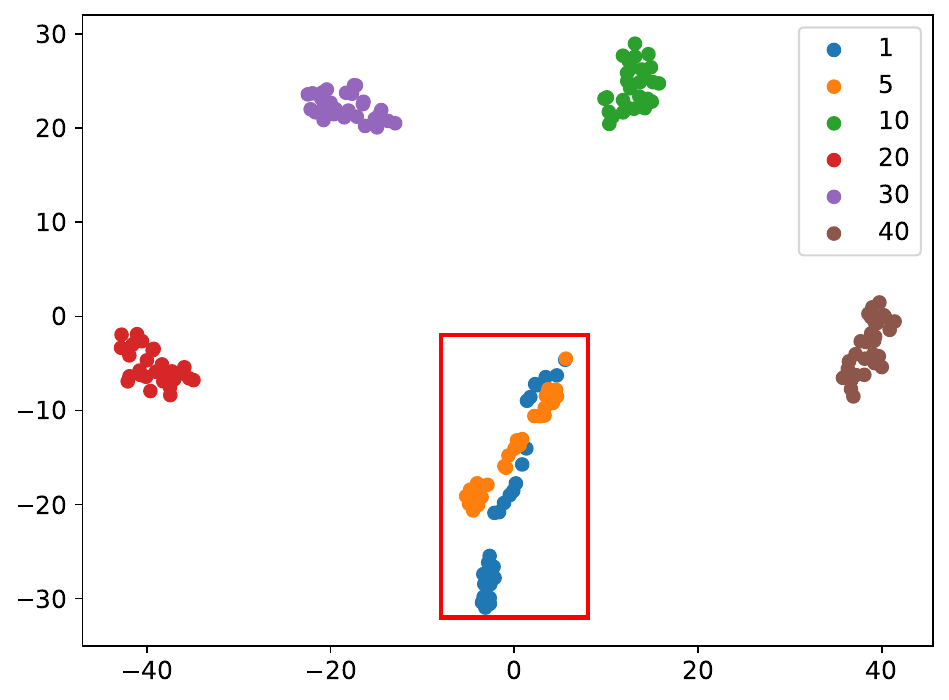} &
            \includegraphics[width=0.25\textwidth]{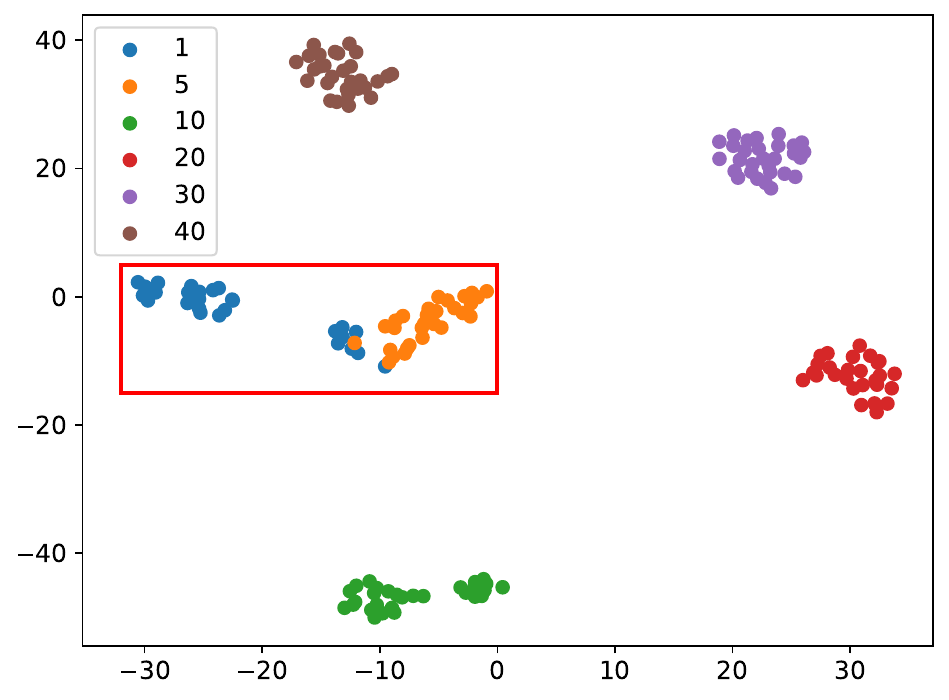} &
            \includegraphics[width=0.25\textwidth]{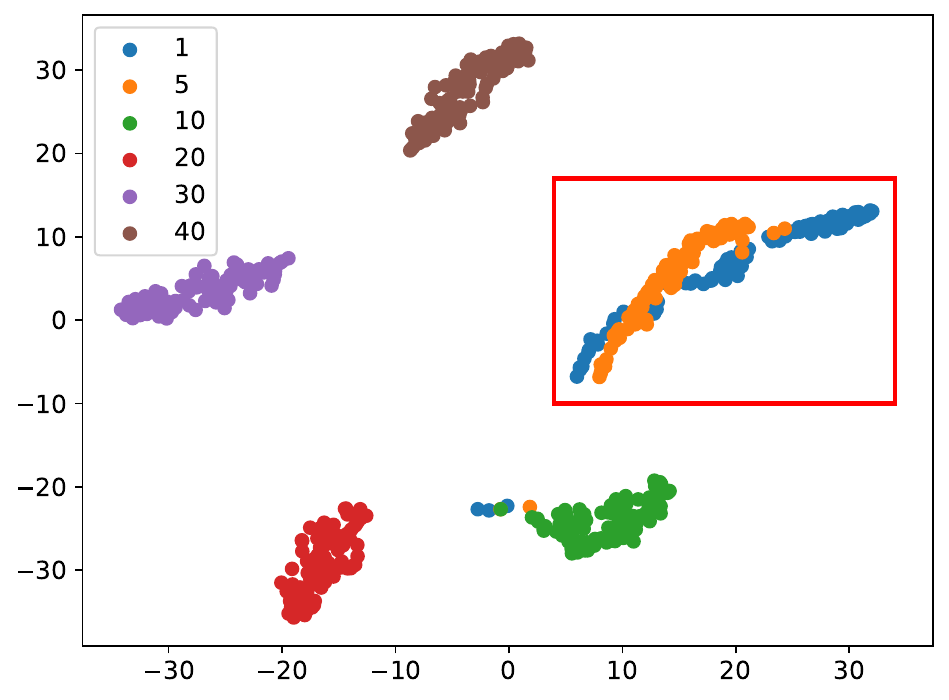}  &
            \includegraphics[width=0.25\textwidth]{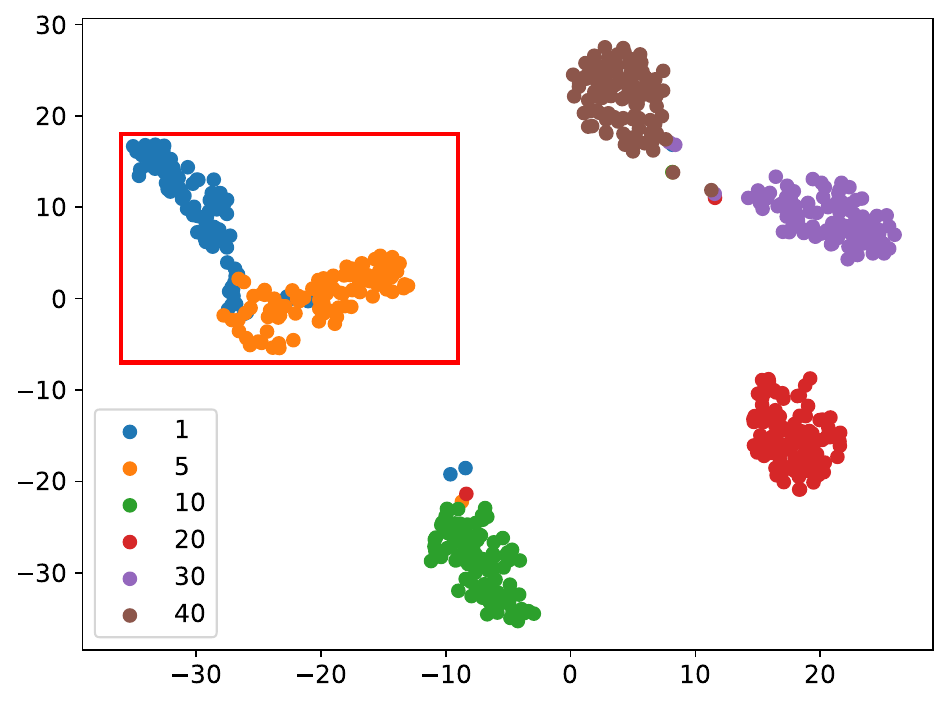}\\
            
           (a) LIVE-1; Explicit learning & (b) LIVE-1; Dual learning & (c) DIV2K-val; Explicit learning & (d) DIV2K-val; Dual learning
        \end{tabular}
    }
    \end{center}
    \vspace{-6.5mm}
    \caption{Visualization of JPEG prior embeddings from CaVE under different training objectives. In (a) and (c), CaVE is trained using only explicit learning, whereas in (b) and (d), it incorporates both explicit and implicit learning. The clusters enclosed in the \textcolor{red}{red box} correspond to unseen compression levels (QF=1,5). Notably, clusters from CaVE with dual learning separate from each other more clearly.}
    \vspace{-6mm}
    \label{fig:clusters}
\end{figure*}

\section{Methodology}
\label{sec:method}
\vspace{-2mm}
\subsection{Preliminaries: One-Step Diffusion Model}
\label{sec:diffusion_models}
\vspace{-2mm}
Latent diffusion models \cite{rombach2022high} operate through a two-stage process: forward diffusion and reverse denoising. During the forward phase, Gaussian noise with variance $\beta_t \in (0, 1)$ is progressively added to the latent representation $\mathbf{z}$: $\mathbf{z}_t = \sqrt{\bar{\alpha}_t} \mathbf{z} + \sqrt{1 - \bar{\alpha}_t} \mathbf{\varepsilon}$, 
where $\mathbf{\varepsilon} \sim \mathcal{N}(0, \mathbf{I})$. Here, the term $\bar{\alpha}_t$ is defined as $\bar{\alpha}_t = \prod_{i=1}^t(1 - \beta_i)$. In the reverse process, the clean latent vector $\hat{z}_0$ is estimated from predicted noise $\hat{\varepsilon}$:
$
\hat{\mathbf{z}}_0 = \frac{\mathbf{z}_t - \sqrt{1 - \bar{\alpha}_t}\hat{\mathbf{\varepsilon}}}{\sqrt{\bar{\alpha}_t}}
$
, where $\mathbf{\hat{\varepsilon}}$ is the prediction of the model $\epsilon_\theta$ given $\mathbf{z_t}$, $t$, and the prior $\mathbf{c}$: $\hat{\varepsilon}=\varepsilon_{\theta}(\mathbf{z}_t; \mathbf{c}, t)$.

Unlike conventional text-to-image (T2I) diffusion models \cite{rombach2022high,podell2023sdxl,saharia2022photorealistic}, \mymodel~feeds the latent representation of the low-quality (LQ) image to the UNet instead of the Gaussian noise. Specifically, our \mymodel~first encodes the LQ image $\mathbf{I}_L$ into the latent space via an encoder $E_\theta$, resulting in $\mathbf{z}_L = E_\theta(\mathbf{I}_L)$. A single-step denoising operation is then performed to predict the noise $\mathbf{\hat{\varepsilon}}$, which allows us to derive the estimated high-quality (HQ) latent representation $\hat{\mathbf{z}}_H$:
\begin{equation}
\hat{\mathbf{z}}_H = \frac{\mathbf{z}_L - \sqrt{1 - \bar{\alpha}_{T_L}} \varepsilon_{\theta} (\mathbf{z}_L; \mathbf{c}_L, T_L)}{\sqrt{\bar{\alpha}_{T_L}}},
\label{eq:generate_zh}
\end{equation}
where $\varepsilon_\theta$ is the denoising network with learnable parameters $\theta$, $\mathbf{c}_L$ is the compression prior, and $T_L\in [0, T]$ is the predefined diffusion timestep. 

Finally, the decoder $D_\theta$ reconstructs HQ image $\hat{\mathbf{I}}_H$ from the latent representation $\hat{\mathbf{z}}_H$: $\hat{\mathbf{I}}_H = D_\theta(\hat{\mathbf{z}}_H).$ Defining the generator $\mathcal{G}_\theta$ and the compression prior $\mathbf{c}_L$, the overall reconstruction from $\mathbf{I}_L$ to $\hat{\mathbf{I}}_H$ can be compactly written as:
\begin{equation}
\hat{\mathbf{I}}_H = \mathcal{G}_\theta(\mathbf{I}_L; \mathbf{c}_L).
\end{equation}
\subsection{Stage 1: Compression-aware Visual Embedder}
\label{sec:cave}
Previous studies have shown that incorporating compression information into the reconstruction process can enhance performance. To incorporate JPEG compression priors into diffusion models, we first train a compression-aware visual embedder (CaVE) to capture these priors effectively. In the second stage, the pre-trained CaVE is leveraged to guide the diffusion model, enabling it to utilize compression-related features from low-quality (LQ) images to improve reconstruction.

\subsubsection{Overall Architecture}
Our CaVE consists of a UNet~\cite{ronneberger2015u} architecture equipped with a lightweight quality factor (QF) predictor designed specifically for extracting compression priors. The UNet decoder and the QF predictor are utilized exclusively for the CaVE training. As illustrated in Fig.~\ref{fig:pipeline}, in explicit learning, the QF predictor guides CaVE to explicitly learn the JPEG compression priors. Simultaneously, implicit learning helps CaVE further capture the relationship between compression and reconstruction, enhancing its ability to model the JPEG compression process.

As illustrated in Fig.~\ref{fig:pipeline}, given a low-quality image $ \mathbf{I}_L \in \mathbb{R}^{H \times W \times 3} $, the CaVE encodes it into a set of feature vectors:
\begin{equation}
    \mathbf{c}_L=\{ \mathbf{c}_{L_k}\in\mathbb{R}^{d} \}_{k=1}^K.
\end{equation}
The QF predictor will then utilize the average feature representation, $\bar{\mathbf{c}}_L=\frac{1}{K}\sum_{k=1}^K \mathbf{c}_{L_k}$, to estimate QF.
Meanwhile, the UNet decoder leverages the extracted feature vectors $\mathbf{c}_L$ to reconstruct the corresponding high-quality image $\hat{\mathbf{I}}_H$.

\subsubsection{Explicit Learning}
\label{subsubsec:explicit}
To explicitly incorporate JPEG compression priors, we train CaVE to produce embeddings that directly facilitate QF prediction~\cite{jiang2021towards,wang2022jpeg}. Specifically, given a low-quality image, we employ the UNet encoder to encode the image into latent representations $\mathbf{c}_L=E(\mathbf{I}_L; \phi)$. The UNet encoder contains four scales, each of which involves residual blocks. Each residual block consists of two 3$\times$3 convolutional layers with Batch Normalization layer and ReLU activation function. 4$\times$4 stride convolutions are employed for downsampling operations. The output dimensions of each scale are set to 64, 128, 256, and 512, respectively. 

We then use the embeddings $\mathbf{z}_L$ from the UNet encoder to explicitly predict the JPEG compression level QF as $\mathit{QF}_{pred}=P(\bar{\mathbf{c}}_L; \phi)$, where $\bar{\mathbf{z}}_L$ is the average of latent embeddings $\mathbf{z}_L$. The QF predictor $P(\cdot;\phi)$ is a lightweight convolutional neural network (CNN) cascaded to a 3-layer multilayer perceptron (MLP). We set the channels in CNN and MLP hidden layers as 512 for a more accurate prediction. 

During training, small patches may contain limited compression-related information (\eg, background regions in compressed images could correspond to multiple QFs). This ambiguity can introduce instability in the training process and hinder CaVE’s ability to effectively capture the underlying JPEG compression patterns~\cite{jiang2021towards}. Therefore, we use the $\mathcal{L}_1$ loss function to mitigate the negative impact of such outliers. Let $B$ be the batch size during training, the QF prediction loss can be expressed as:
\begin{equation}
    \mathcal{L}_{\mathit{QF}}=\frac{1}{B}\sum_{i=1}^B\| \mathit{QF}_{pred}^i-\mathit{QF}_{gt}^i\|_1.
\end{equation}
To illustrate that our CaVE learns discriminative representations for images with different QFs, we visualize $\mathbf{z_L}$ from the UNet encoder using the t-SNE~\cite{van2008visualizing} method, as shown in Fig.~\ref{fig:clusters}. Our results show that CaVE, when trained with the QF loss, effectively distinguishes compression qualities present in the training set. However, it struggles to differentiate unseen compression levels (\eg, QF = 1, 5), indicating that relying solely on the QF loss fails to capture the full spectrum of compression-related information.

\subsubsection{Implicit Learning}
To more comprehensively capture JPEG compression priors, we train CaVE to produce embeddings that not only predict the quality factor (QF) (explicit learning) but also reconstruct high-quality images (implicit learning). By incorporating a reconstruction objective, CaVE implicitly learns the intricate details of the JPEG compression process.

Specifically, given a low-quality image $\mathbf{I}_L$, we employ the UNet-based architecture~\cite{ronneberger2015u} to restore the corresponding high-quality image: $\hat{\mathbf{I}}_{H}= D(\mathbf{z}_L; \phi)$. We use the same encoder in Sec.~\ref{subsubsec:explicit}. The decoder, on the other hand, includes three scales, each of which contains residual blocks that consist of two 3$\times$3 convolutional layers with Batch Normalization layer and ReLU activation function. Each scale receives the corresponding intermediate image features from the encoder to generate the restored image. 

Given a batch of $B$ training samples, the objective of implicit learning is to minimize the $\mathcal{L}_1$ loss between the reconstructed image $\hat{\mathbf{I}}_{H}$ and the ground-truth $\mathbf{I}_{H}$:
\begin{equation}
    \mathcal{L}_{\mathit{rec}}=\frac{1}{B}\sum_{i=1}^B\| \hat{\mathbf{I}}_{H}^i-\mathbf{I}_{H}^i\|_1.
\end{equation}
The overall CaVE training objective is formulated as:
\begin{equation}
    \mathcal{L}_{\mathit{CaVE}}=\mathcal{L}_{\mathit{QF}}+\lambda \cdot \mathcal{L}_{\mathit{rec}}.
\end{equation}

To demonstrate how implicit learning enhances CaVE’s understanding of JPEG compression, we visualize its learned embeddings in Fig.~\ref{fig:clusters}. Our results reveal that the dual learning allows CaVE to capture more nuanced variations in compression artifact, enabling it to differentiate even previously unseen compression levels. 

\subsection{Stage 2: JPEG Artifact Removal}
\label{sec:diff}
After extracting compression prior embeddings, we integrate them into our OSD model and fine-tune the pre-trained diffusion model using LoRA~\cite{hu2021lora}. The overall training objective combines perceptual and GAN losses. Specifically, perceptual loss directly aligns with the reconstruction process to ensure high-fidelity restoration. The GAN loss~\cite{goodfellow2020generative} enhances the realism of the generated images by refining textures and preserving natural image characteristics. By jointly optimizing these loss functions, our model effectively achieve high-quality, photorealistic reconstructions.

\noindent\textbf{Perceptual Loss.} While $\mathcal{L}_2$ loss is widely used for image reconstruction due to its simplicity and pixel-wise precision, it often struggles to capture perceptual quality. Minimizing $\mathcal{L}_2$ loss tends to produce overly smooth outputs, as it treats all pixel differences equally, ignoring structural and textural information critical for human perception.

To address this issue, we additionally incorporate DISTS (Deep Image Structure and Texture Similarity) loss~\cite{ding2020image}, which goes beyond pixel-wise comparisons by measuring perceptual similarity between images. Unlike $\mathcal{L}_2$ loss, DISTS is designed to align with human visual perception, capturing both structural details and textural characteristics. The perceptual loss can be written as:
\begin{equation}
    \mathcal{L}_{per}=\mathcal{L}_2(\hat{\mathbf{I}}_H, \mathbf{I}_H) + \lambda_{D} \mathcal{L}_{DISTS}(\hat{\mathbf{I}}_H, \mathbf{I}_H).
\end{equation}

\noindent\textbf{GAN Loss.} Generating stable images remains a significant challenge for OSD models, primarily due to their constrained computational resources. Prior research~\cite{yin2024dmd,yin2024dmd2,wang2024sinsr,wu2024one} has largely relied on distillation techniques to transfer knowledge from multi-step diffusion (MSD) models. However, these approaches are inherently limited by the performance of their teacher models, restricting their potential for further improvement. 

To overcome these limitations, we adopt an alternative strategy by integrating a discriminative network to enhance the realism of restoration. The GAN losses~\cite{goodfellow2020generative} used to train the generator $G_\theta$ and discriminator $D_\theta$ are defined as:
\begin{equation}
    \mathcal{L}_\mathcal{G} = -\mathbb{E}_t \left[\log \mathcal{D}_{\theta} \left( \hat{\mathbf{z}}_H\right)\right],
\end{equation}
\begin{equation}
\begin{aligned}
\mathcal{L}_\mathcal{D} = &-\mathbb{E}_t \left[\log \left( 1 - \mathcal{D}_{\theta} \left( \hat{\mathbf{z}}_H \right) \right)\right] \\
&- \mathbb{E}_t \left[\log \mathcal{D}_{\theta} \left(\mathbf{z}_H \right)\right], 
\end{aligned}
\end{equation}
where $\hat{\mathbf{z}}_H$ and $\mathbf{z}_H$ are the latent embeddings of the reconstructed image and the ground-truth image, respectively.

In our JPEG artifact removal pipeline, we first train CaVE in stage 1, then we incorporate the perceptual and GAN losses to fine-tune our OSD with JPEG compression priors from CaVE. The overall training objective for the generator is formulated as:
\begin{equation}
    \mathcal{L}=\mathcal{L}_{per} + \lambda_{G} \mathcal{L}_\mathcal{G}.
\end{equation}
\section{Experiments}
\subsection{Experimental Settings}
\noindent\textbf{Training and Testing Datasets.} Following the previous work~\cite{wu2024one,li2024promptcir}, We use DF2K which comprises 800 images from DIV2K~\cite{agustsson2017ntire} and 2,650 images from Flickr2K~\cite{timofte2017ntire}, and LSDIR~\cite{li2023lsdir} as our training datasets, totaling 88,441 different high-quality images. During training, we randomly crop 256$\times$256 patches and dynamically synthesize low-quality and high-quality image pairs by sampling the quality factor from a range of 8 to 95. We evaluate \mymodel~and compare its performance against competing methods using LIVE-1~\cite{sheikh2005live}, Urban100~\cite{huang2015single}, and DIV2K-Val~\cite{agustsson2017ntire} datasets.

\noindent\textbf{Evaluation Metrics.}
To ensure a comprehensive and holistic evaluation of different methods, we employ a diverse set of both full-reference and no-reference image quality metrics. For perceptual quality assessment with a reference image, we utilize LPIPS~\cite{zhang2018unreasonable} and DISTS~\cite{ding2020image}, both of which measure structural and perceptual similarity by leveraging deep feature representations. In addition, to assess image quality without requiring a reference, we adopt MANIQA~\cite{yang2022maniqa}, MUSIQ~\cite{ke2021musiq}, and CLIPIQA~\cite{wang2023exploring}, which are designed to predict human-perceived image quality. 

\begin{table*}[t]
\scriptsize
\setlength{\tabcolsep}{0.4mm}

\newcolumntype{?}{!{\vrule width 1pt}}
\newcolumntype{C}{>{\centering\arraybackslash}X}

\begin{center}

\begin{tabularx}{\textwidth}{l|*{5}{C}|*{5}{C}|*{5}{C}}
\toprule[0.15em]
\rowcolor{gray} & \multicolumn{5}{c|}{QF=5} & \multicolumn{5}{c|}{QF=10} & \multicolumn{5}{c}{QF=20} \\
\rowcolor{gray}
\multirow{-2}{*}{Methods}  & LPIPS$\downarrow$ & DISTS$\downarrow$ & MUSIQ$\uparrow$ & M-IQA$\uparrow$& C-IQA$\uparrow$ & LPIPS$\downarrow$ & DISTS$\downarrow$ & MUSIQ$\uparrow$ & M-IQA$\uparrow$ & C-IQA$\uparrow$ & LPIPS$\downarrow$ & DISTS$\downarrow$ & MUSIQ$\uparrow$ & M-IQA$\uparrow$ & C-IQA$\uparrow$ \\
\midrule[0.15em]
JPEG~\cite{wallace1991jpeg} & 0.4384 & 0.3242 & 40.33 & 0.2294 & 0.1716 & 0.3013 & 0.2387 & 53.88 & 0.3509 & 0.2737 & 0.1799 & 0.1653 & 64.12 & 0.4411 & 0.5542 \\
FBCNN~\cite{jiang2021towards} & 0.3736 & 0.2353 & 63.56 & \textcolor{iccvblue}{0.3425} & 0.2763 & 0.2503 & 0.1785 & 71.00 & 0.4207 & 0.4767 & 0.1583 & 0.1319 & 73.96 & 0.4551 & 0.5535 \\
JDEC~\cite{han2024jdec} & 0.4113 & 0.2364 & 55.66 & 0.2002 & 0.1539 & 0.2450 & 0.1740 & 70.80 & 0.4065 & 0.4811 & 0.1555 & 0.1282 & 73.81 & 0.4433 & 0.5512 \\
PromptCIR~\cite{li2024promptcir} & 0.3797 & 0.2334 & 60.34 & 0.2790 & 0.2655 & 0.2290 & 0.1658 & \textcolor{iccvblue}{72.39} & 0.4500 & 0.5176 & 0.1450 & 0.1223 & \textcolor{iccvblue}{74.12} & 0.4713 & 0.5847 \\
\midrule
DiffBIR*~\cite{lin2024diffbir} (s=50) & 0.3509 & 0.2035 & 58.09 & 0.2812 & 0.3776 & 0.2160 & 0.1319 & 67.38 & 0.3789 & 0.5789 & 0.1500 & 0.0988 & 71.08 & 0.4371 & 0.6814 \\
SUPIR~\cite{yu2024scaling} (s=50) & 0.4856 & 0.2720 & 52.69 & 0.3229 & 0.3149 & 0.2770 & 0.1558 & 68.77 & \textcolor{iccvblue}{0.5183} & 0.6115 & 0.1683 & 0.1121 & 73.02 & \textcolor{red}{0.6237} & \textcolor{iccvblue}{0.7364} \\
OSEDiff*~\cite{wu2024one} (s=1) & \textcolor{iccvblue}{0.2675} & \textcolor{iccvblue}{0.1653} & \textcolor{iccvblue}{65.51} & 0.3417 & \textcolor{iccvblue}{0.5623} & \textcolor{iccvblue}{0.1749} & \textcolor{iccvblue}{0.1164} & 71.23 & 0.3963 & \textcolor{iccvblue}{0.7022} & \textcolor{iccvblue}{0.1270} & \textcolor{iccvblue}{0.0856} & 72.70 & 0.4219 & 0.7260 \\
\textbf{\mymodel}~(ours, s=1) & \textcolor{red}{0.2062} & \textcolor{red}{0.1121} & \textcolor{red}{73.16} & \textcolor{red}{0.5321} & \textcolor{red}{0.7212} & \textcolor{red}{0.1428} & \textcolor{red}{0.0867} & \textcolor{red}{74.39} & \textcolor{red}{0.5438} & \textcolor{red}{0.7559} & \textcolor{red}{0.1101} & \textcolor{red}{0.0692} & \textcolor{red}{74.34} & \textcolor{iccvblue}{0.5323} & \textcolor{red}{0.7565} \\

\bottomrule[0.15em]
\end{tabularx}

\vspace{1mm}
\small
(a) Quantitative comparison on the LIVE-1 dataset
\vspace{1mm}
\scriptsize

\begin{tabularx}{\textwidth}{l|*{5}{C}|*{5}{C}|*{5}{C}}
\toprule[0.15em]
\rowcolor{gray} & \multicolumn{5}{c|}{QF=5} & \multicolumn{5}{c|}{QF=10} & \multicolumn{5}{c}{QF=20} \\
\rowcolor{gray}
\multirow{-2}{*}{Methods}  & LPIPS$\downarrow$ & DISTS$\downarrow$ & MUSIQ$\uparrow$ & M-IQA$\uparrow$& C-IQA$\uparrow$ & LPIPS$\downarrow$ & DISTS$\downarrow$ & MUSIQ$\uparrow$ & M-IQA$\uparrow$ & C-IQA$\uparrow$ & LPIPS$\downarrow$ & DISTS$\downarrow$ & MUSIQ$\uparrow$ & M-IQA$\uparrow$ & C-IQA$\uparrow$ \\
\midrule[0.15em]
JPEG~\cite{wallace1991jpeg} & 0.3481 & 0.2834 & 50.46 & 0.3656 & 0.2806 & 0.2254 & 0.2145 & 60.87 & 0.4401 & 0.3517 & 0.1244 & 0.1521 & 67.60 & 0.4967 & 0.5343 \\
FBCNN~\cite{jiang2021towards} & 0.2341 & 0.2162 & 69.03 & 0.4263 & 0.3800 & 0.1462 & 0.1648 & 72.55 & 0.5033 & 0.5014 & 0.0896 & 0.1249 & \textcolor{iccvblue}{73.39} & 0.5288 & 0.5437 \\
JDEC~\cite{han2024jdec} & 0.2794 & 0.2309 & 62.97 & 0.3386 & 0.2518 & 0.1382 & 0.1570 & 72.52 & 0.5001 & 0.4959 & 0.0846 & 0.1175 & 73.30 & 0.5230 & 0.5369 \\
PromptCIR~\cite{li2024promptcir} & 0.2389 & 0.2037 & 66.08 & 0.3946 & 0.3619 & \textcolor{iccvblue}{0.1183} & 0.1431 & \textcolor{red}{73.01} & 0.5380 & 0.5337 & \textcolor{iccvblue}{0.0739} & 0.1083 & \textcolor{red}{73.47} & 0.5489 & 0.5662 \\
\midrule
DiffBIR*~\cite{lin2024diffbir} (s=50) & 0.2018 & \textcolor{iccvblue}{0.1657} & 69.63 & 0.4285 & 0.5470 & 0.1344 & 0.1207 & 71.77 & 0.4813 & 0.5966 & 0.1005 & 0.0939 & 72.51 & 0.5105 & 0.6306 \\
SUPIR~\cite{yu2024scaling} (s=50) & 0.3279 & 0.2018 & \textcolor{iccvblue}{69.94} & \textcolor{iccvblue}{0.5546} & 0.5536 & 0.2489 & 0.1659 & 72.57 & \textcolor{red}{0.5995} & 0.6178 & 0.2125 & 0.1518 & 73.01 & \textcolor{red}{0.6105} & 0.6397 \\
OSEDiff*~\cite{wu2024one} (s=1) & \textcolor{iccvblue}{0.1959} & 0.1690 & 68.60 & 0.4491 & \textcolor{iccvblue}{0.5710} & 0.1262 & \textcolor{iccvblue}{0.1168} & 71.55 & 0.4927 & \textcolor{iccvblue}{0.6364} & 0.0911 & \textcolor{iccvblue}{0.0860} & 72.43 & 0.5082 & \textcolor{iccvblue}{0.6591} \\
\textbf{\mymodel}~(ours, s=1) & \textcolor{red}{0.1407} & \textcolor{red}{0.1101} & \textcolor{red}{72.16} & \textcolor{red}{0.5693} & \textcolor{red}{0.6741} & \textcolor{red}{0.0974} & \textcolor{red}{0.0842} & \textcolor{iccvblue}{72.61} & \textcolor{iccvblue}{0.5725} & \textcolor{red}{0.6824} & \textcolor{red}{0.0753} & \textcolor{red}{0.0667} & 72.63 & \textcolor{iccvblue}{0.5694} & \textcolor{red}{0.6830} \\

\bottomrule[0.15em]
\end{tabularx}

\vspace{1mm}
\small
(b) Quantitative comparison on the Urban100 dataset
\vspace{1mm}
\scriptsize

\begin{tabularx}{\textwidth}{l|*{5}{C}|*{5}{C}|*{5}{C}}
\toprule[0.15em]
\rowcolor{gray} & \multicolumn{5}{c|}{QF=5} & \multicolumn{5}{c|}{QF=10} & \multicolumn{5}{c}{QF=20} \\
\rowcolor{gray}
\multirow{-2}{*}{Methods}  & LPIPS$\downarrow$ & DISTS$\downarrow$ & MUSIQ$\uparrow$ & M-IQA$\uparrow$& C-IQA$\uparrow$ & LPIPS$\downarrow$ & DISTS$\downarrow$ & MUSIQ$\uparrow$ & M-IQA$\uparrow$ & C-IQA$\uparrow$ & LPIPS$\downarrow$ & DISTS$\downarrow$ & MUSIQ$\uparrow$ & M-IQA$\uparrow$ & C-IQA$\uparrow$ \\
\midrule[0.15em]
JPEG~\cite{wallace1991jpeg} & 0.4466 & 0.3183 & 34.59 & 0.2570 & 0.2595 & 0.3234 & 0.2255 & 47.53 & 0.3120 & 0.3303 & 0.2072 & 0.1465 & 57.45 & 0.3557 & 0.5072 \\
FBCNN~\cite{jiang2021towards} & 0.3445 & 0.2078 & 56.52 & 0.3025 & 0.3004 & 0.2448 & 0.1581 & 61.79 & 0.3593 & 0.4561 & 0.1733 & 0.1168 & 65.20 & 0.3775 & 0.5221 \\
JDEC~\cite{han2024jdec} & 0.3811 & 0.2234 & 53.88 & 0.2118 & 0.1841 & 0.2313 & 0.1574 & \textcolor{red}{67.48} & 0.3689 & 0.4675 & 0.1565 & 0.1152 & \textcolor{red}{69.90} & 0.3927 & 0.5319 \\
PromptCIR~\cite{li2024promptcir} & 0.3549 & 0.2067 & 52.21 & 0.2705 & 0.3041 & 0.2240 & 0.1459 & 62.63 & \textcolor{iccvblue}{0.3758} & 0.4956 & 0.1581 & 0.1061 & 65.62 & 0.3871 & 0.5483 \\
\midrule
DiffBIR*~\cite{lin2024diffbir} (s=50) & 0.2788 & 0.1533 & 60.21 & 0.3220 & \textcolor{iccvblue}{0.4975} & 0.1953 & 0.1072 & 65.22 & 0.3754 & 0.5912 & 0.1542 & 0.0856 & \textcolor{iccvblue}{67.06} & \textcolor{iccvblue}{0.4033} & \textcolor{iccvblue}{0.6355} \\
SUPIR~\cite{yu2024scaling} (s=50) & 0.4372 & 0.2148 & 54.07 & \textcolor{iccvblue}{0.3438} & 0.4219 & 0.3121 & 0.1410 & 61.93 & 0.3570 & 0.5186 & 0.2295 & 0.1161 & 64.87 & 0.3723 & 0.5535 \\
OSEDiff*~\cite{wu2024one} (s=1) & \textcolor{iccvblue}{0.2624} & \textcolor{iccvblue}{0.1474} & \textcolor{iccvblue}{60.83} & 0.3252 & 0.4974 & \textcolor{iccvblue}{0.1823} & \textcolor{iccvblue}{0.0996} & 64.88 & 0.3640 & \textcolor{iccvblue}{0.6208} & \textcolor{iccvblue}{0.1341} & \textcolor{iccvblue}{0.0721} & 66.10 & 0.3689 & 0.6323 \\
\textbf{\mymodel}~(ours, s=1) & \textcolor{red}{0.2086} & \textcolor{red}{0.0994} & \textcolor{red}{66.28} & \textcolor{red}{0.4069} & \textcolor{red}{0.6413} & \textcolor{red}{0.1436} & \textcolor{red}{0.0714} &\textcolor{iccvblue}{66.97} & \textcolor{red}{0.4113} & \textcolor{red}{0.6498} & \textcolor{red}{0.1020} & \textcolor{red}{0.0511} & 66.66 & \textcolor{red}{0.4042} & \textcolor{red}{0.6443} \\

\bottomrule[0.15em]
\end{tabularx}

\vspace{1mm}
\small
(c) Quantitative comparison on the DIV2K-val dataset

\vspace{-2.5mm}
\caption{Quantitative comparison on LIVE-1, Urban100, and DIV2K-Val datasets with non-diffusion, MSD, and OSD methods. M-IQA stands for MANIQA, and C-IQA stands for CLIPIQA. The best and second best results are colored with \textcolor{red}{red} and \textcolor{iccvblue}{blue}. DiffBIR* and OSEDiff* are retrained on the same dataset as our method for reference.}
\label{table:main_results}
\end{center}
\vspace{-8mm}
\end{table*}

\noindent\textbf{Implementation Details.} We use PyTorch~\citep{paszke2019pytorch} to implement our model. In the first stage (Sec.~\ref{sec:cave}), we optimize CaVE using the Adam optimizer~\cite{kingma2014adam} with a learning rate of 2$\times$10$^{-5}$ and a batch size of $B$=128. The weight for implicit compression prior learning, $\lambda$, is set to 1,000. We train CaVE for 200K iterations with 4 NVIDIA A6000 GPUs.

In the second stage (Sec.~\ref{sec:diff}), we train \mymodel~using the AdamW optimizer~\cite{loshchilov2017decoupled} with a learning rate of 5$\times$10$^{-5}$, weight decay of 1$\times$10$^{-5}$, and a batch size of 32 for both the generator and discriminator. The discriminator employs a pre-trained StableDiffusion UNet encoder and a lightweight MLP. We set the LoRA~\cite{hu2021lora} rank to 16 for fine-tuning. We utilize CaVE to construct the prompt embeddings. The discriminator follows the same training setup as \mymodel. The weight for DISTS loss $\lambda_{D}$ and GAN loss $\lambda_{G}$ is set to 1 and 5$\times$10$^{-3}$, respectively. This phase is trained for 100K iterations on 4 NVIDIA A6000 GPUs. 

\begin{figure*}[t]
\scriptsize
\centering
\begin{tabular}{cc}

\resizebox{1.\textwidth}{!}{
    \begin{tabular}{c}
    
    \hspace{-0.4cm}
    \includegraphics[width=0.2\textwidth, height=0.174\textwidth]{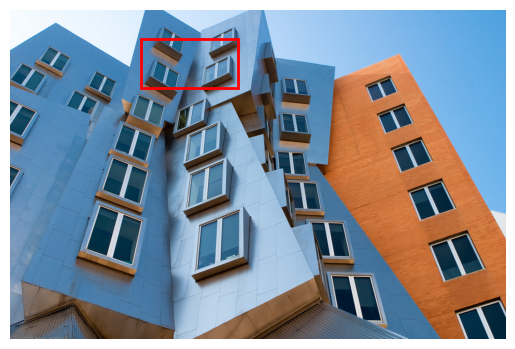}
    \\
    Urban100: img\_091
    \end{tabular}
    \hspace{-0.6cm}
    \begin{tabular}{cccc}
    \includegraphics[width=0.143\textwidth]{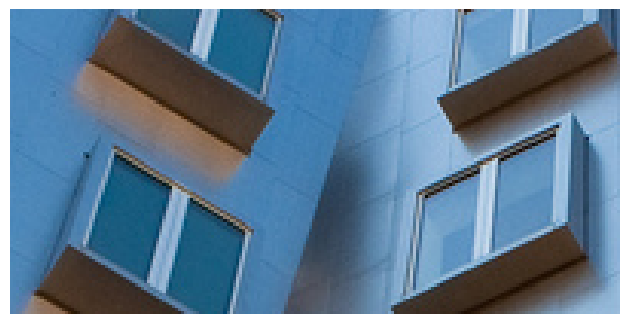} \hspace{-5mm} &
    \includegraphics[width=0.143\textwidth]{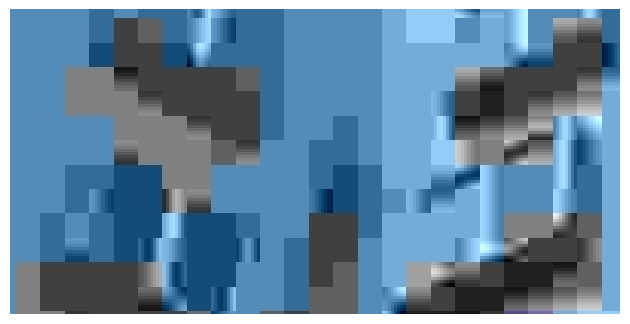} \hspace{-5mm} &
    \includegraphics[width=0.143\textwidth]{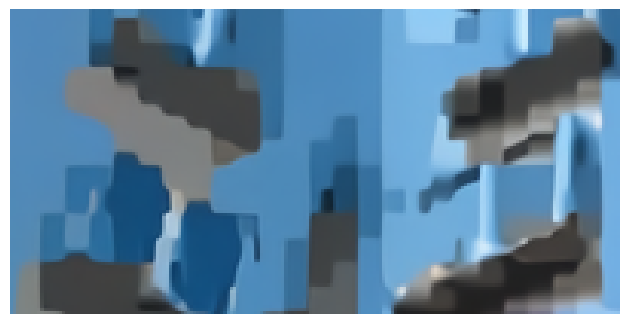} \hspace{-5mm} &
    \includegraphics[width=0.143\textwidth]{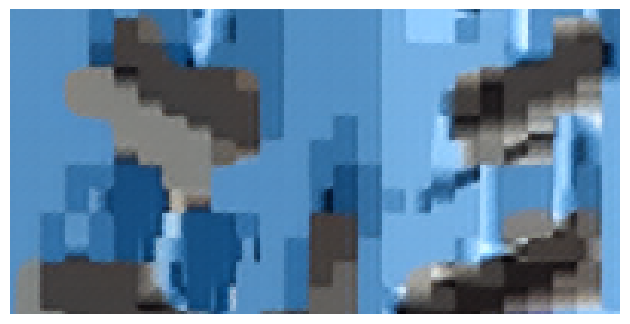}
    \\
    HQ \hspace{-5mm} &
    JPEG (QF=1) \hspace{-5mm} &
    FBCNN~\cite{jiang2021towards} \hspace{-5mm} &
    JDEC~\cite{han2024jdec}
    \\
    \includegraphics[width=0.143\textwidth]{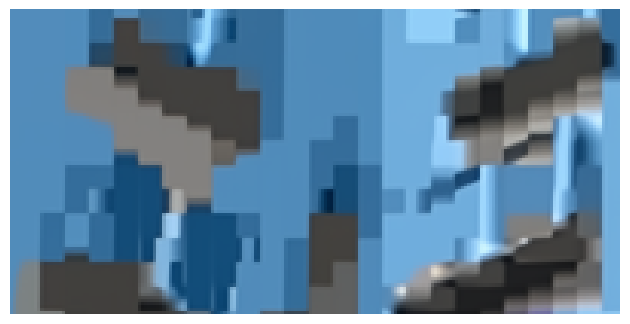} \hspace{-5mm} &
    \includegraphics[width=0.143\textwidth]{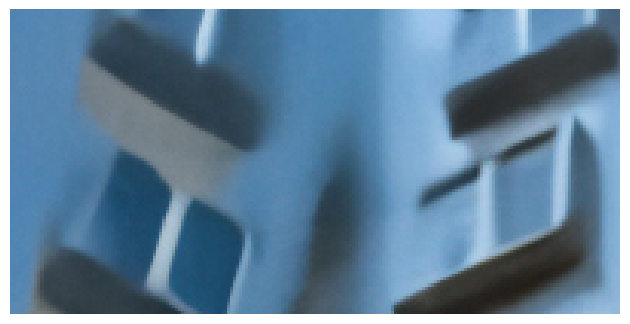} \hspace{-5mm} &
    \includegraphics[width=0.143\textwidth]{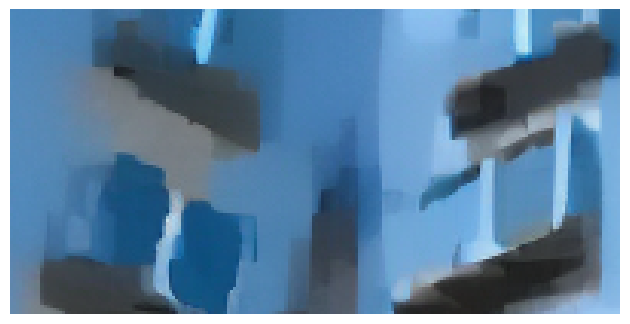} \hspace{-5mm} &
    \includegraphics[width=0.143\textwidth]{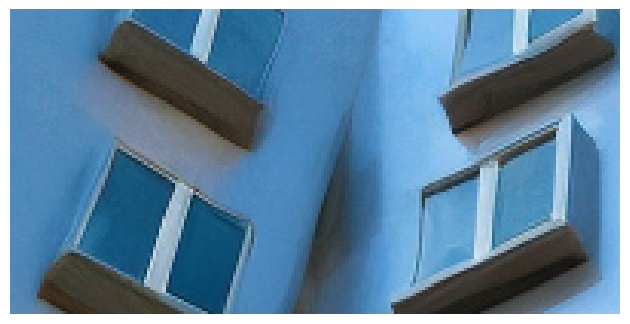}  
    \\ 
    PromptCIR~\cite{li2024promptcir} \hspace{-5mm} &
    DiffBIR*~\cite{lin2024diffbir} \hspace{-5mm} &
    OSEDiff*~\cite{wu2024one} \hspace{-5mm} &
    CODiff (ours)
    \\
    \end{tabular}
}
\\

\resizebox{1.\textwidth}{!}{
    \begin{tabular}{c}
    
    \hspace{-0.4cm}
    \includegraphics[width=0.2\textwidth]{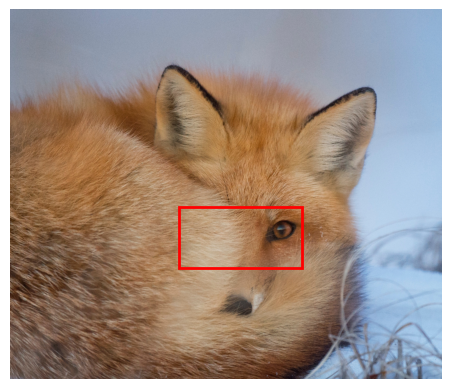}
    \\
    DIV2K-Val: 0862
    \end{tabular}
    
    \hspace{-0.6cm}
    \begin{tabular}{cccc}
    \includegraphics[width=0.143\textwidth]{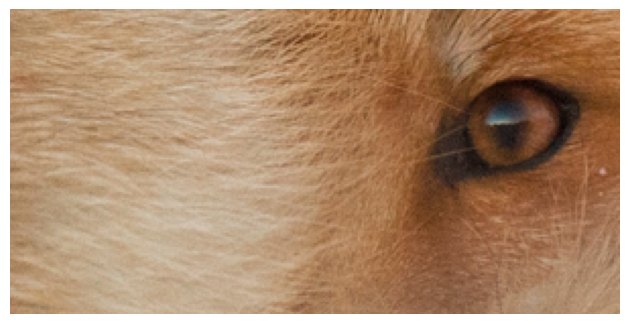} \hspace{-5mm} &
    \includegraphics[width=0.143\textwidth]{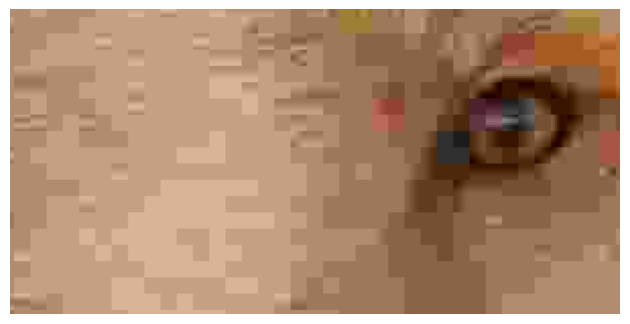} \hspace{-5mm} &
    \includegraphics[width=0.143\textwidth]{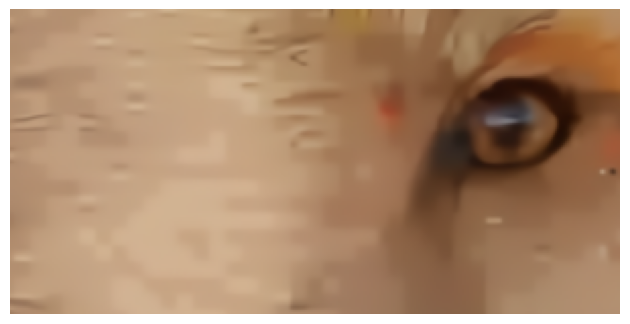} \hspace{-5mm} &
    \includegraphics[width=0.143\textwidth]{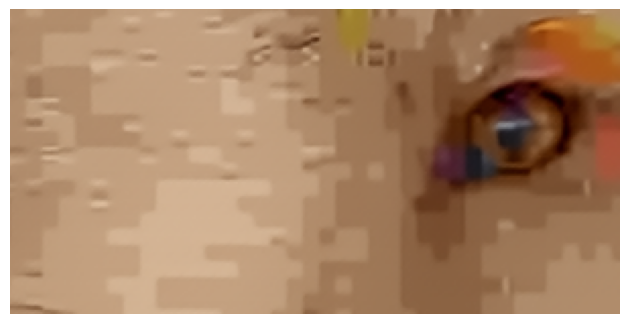}
    \\
    HQ \hspace{-5mm} &
    JPEG (QF=5) \hspace{-5mm} &
    FBCNN~\cite{jiang2021towards} \hspace{-5mm} &
    JDEC~\cite{han2024jdec}
    \\
    \includegraphics[width=0.143\textwidth]{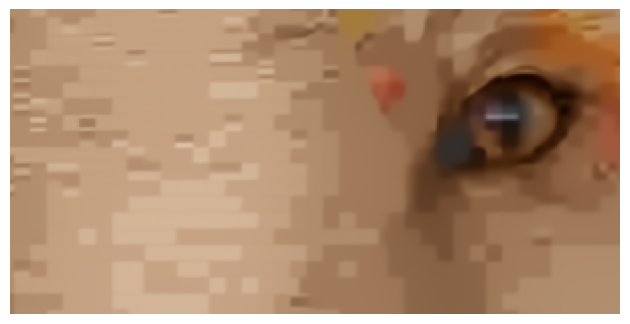} \hspace{-5mm} &
    \includegraphics[width=0.143\textwidth]{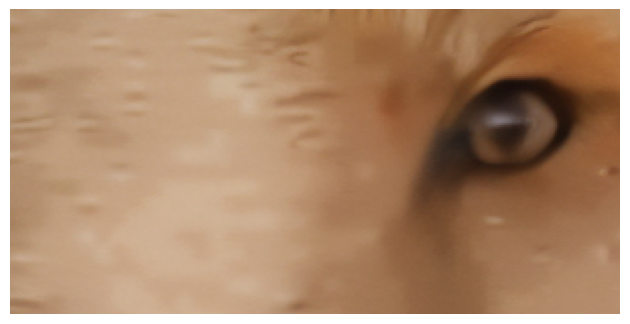} \hspace{-5mm} &
    \includegraphics[width=0.143\textwidth]{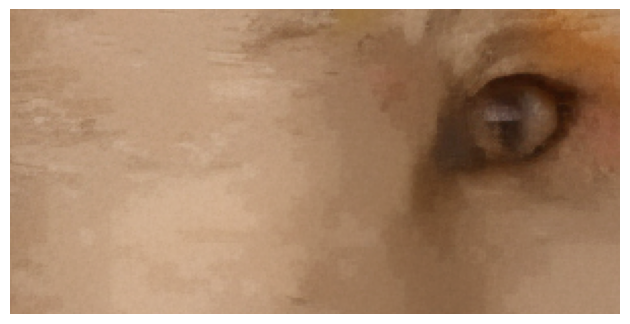} \hspace{-5mm} &
    \includegraphics[width=0.143\textwidth]{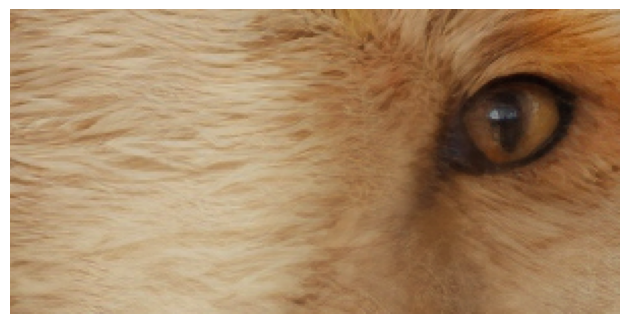}  
    \\ 
    PromptCIR~\cite{li2024promptcir} \hspace{-5mm} &
    DiffBIR*~\cite{lin2024diffbir} \hspace{-5mm} &
    OSEDiff*~\cite{wu2024one} \hspace{-5mm} &
    \mymodel~(ours)
    \\
    \end{tabular}
} \\

\resizebox{1.\textwidth}{!}{
    \begin{tabular}{c}
    
    \hspace{-0.4cm}
    \includegraphics[width=0.2\textwidth, height=0.171\textwidth]{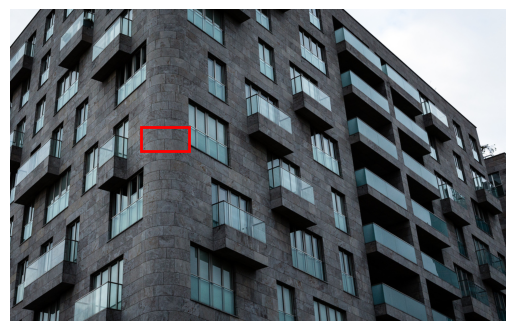}
    \\
    Urban100: img\_080
    \end{tabular}
    \hspace{-0.6cm}
    \begin{tabular}{cccc}
    \includegraphics[width=0.143\textwidth]{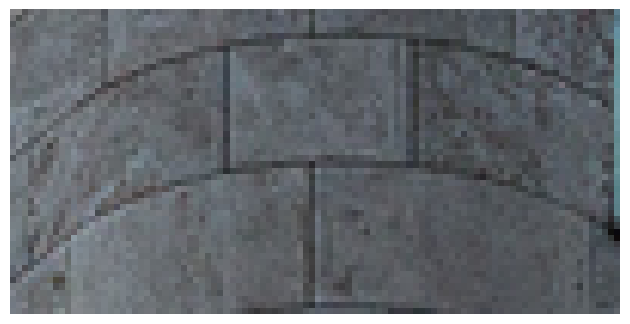} \hspace{-5mm} &
    \includegraphics[width=0.143\textwidth]{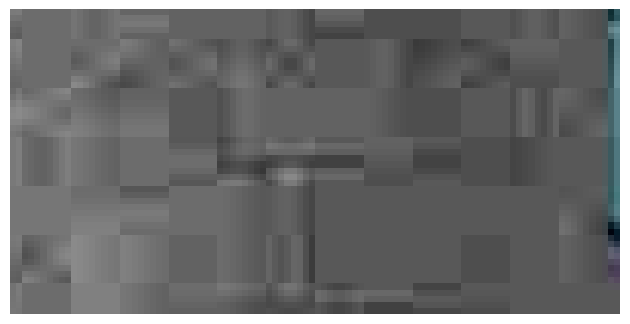} \hspace{-5mm} &
    \includegraphics[width=0.143\textwidth]{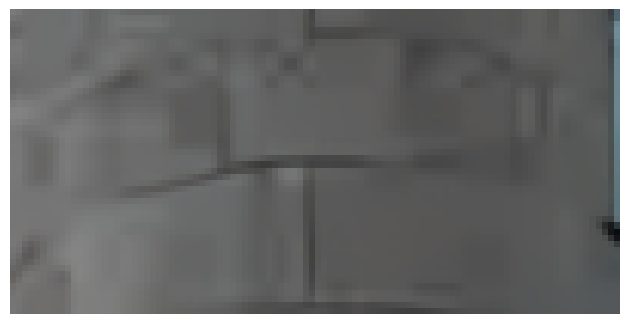} \hspace{-5mm} &
    \includegraphics[width=0.143\textwidth]{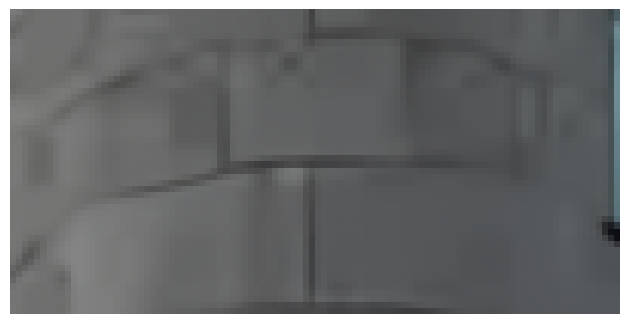}
    \\
    HQ \hspace{-5mm} &
    JPEG (QF=10) \hspace{-5mm} &
    FBCNN~\cite{jiang2021towards} \hspace{-5mm} &
    JDEC~\cite{han2024jdec}
    \\
    \includegraphics[width=0.143\textwidth]{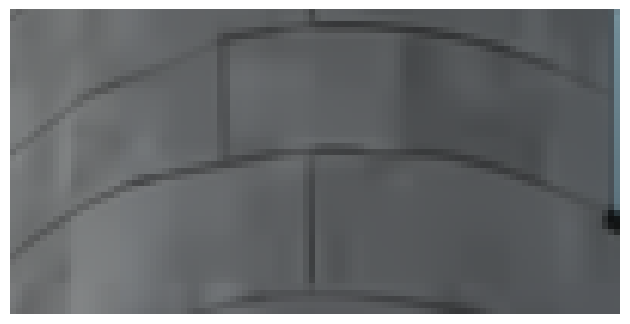} \hspace{-5mm} &
    \includegraphics[width=0.143\textwidth]{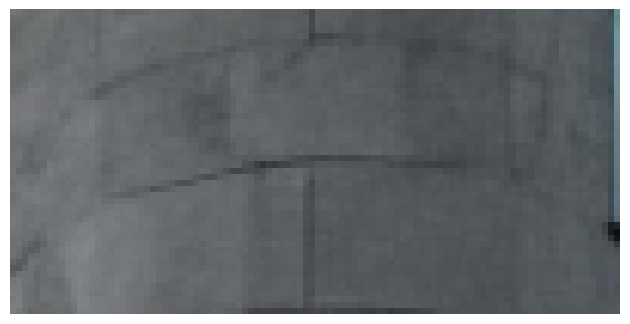} \hspace{-5mm} &
    \includegraphics[width=0.143\textwidth]{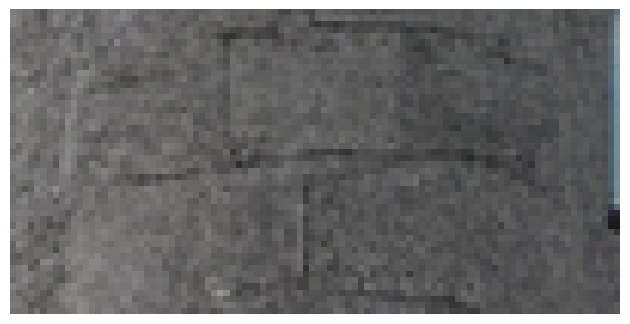} \hspace{-5mm} &
    \includegraphics[width=0.143\textwidth]{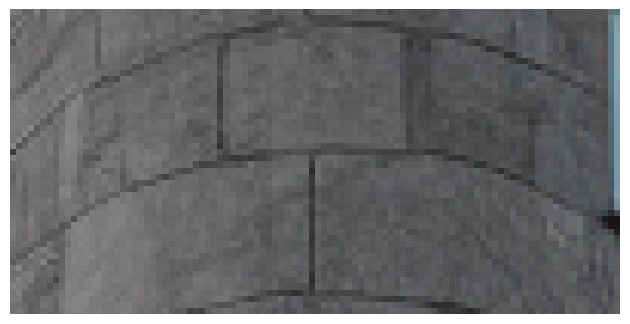}  
    \\ 
    PromptCIR~\cite{li2024promptcir} \hspace{-5mm} &
    DiffBIR*~\cite{lin2024diffbir} \hspace{-5mm} &
    OSEDiff*~\cite{wu2024one} \hspace{-5mm} &
    CODiff (ours)
    \\
    \end{tabular}
}

\end{tabular}
\vspace{-4mm}
\caption{Visual comparison on different quality factors (QF=1, 5 and 10). DiffBIR* and OSEDiff* are retrained as references.}
\label{fig:main_comp}
\vspace{-5mm}
\end{figure*}

\subsection{Comparison with SOTA Methods}
\noindent\textbf{Compared Methods.} We compare \mymodel~with state-of-the-art CNN and Transformer-based methods specifically designed for JPEG artifact removal, including FBCNN~\cite{jiang2021towards}, JDEC~\cite{han2024jdec}, and PromptCIR~\cite{li2024promptcir}. Additionally, we evaluate \mymodel~against recent leading diffusion-based image restoration methods, including DiffBIR~\cite{lin2024diffbir}, SUPIR~\cite{yu2024scaling}, and OSEDiff~\cite{wu2024one}. Among them, DiffBIR and OSEDiff are built upon the pre-trained StableDiffusion~\cite{rombach2022high} model. While SUPIR leverages StableDiffusion-XL~\cite{podell2023sdxl}, a larger model with 2.6 billion parameters. We re-train DiffBIR and OSEDiff under the same experimental settings as ours and denote them as DiffBIR* and OSEDiff* respectively.

\noindent\textbf{Quantitative Results.} 
The quantitative comparisons on LIVE-1, Urban100, and DIV2K-Val datasets are summarized in Tab.~\ref{table:main_results}. \mymodel~consistently outperforms competing methods across a diverse range of evaluation metrics. Specifically, it achieves notable improvements in both full-reference metrics (LPIPS and DISTS) and no-reference metrics (MUSIQ, MANIQA, and CLIPIQA). Those comparisons demonstrate its superior ability to recover visually pleasing and perceptually accurate high-quality images.

Compared to existing OSD models, \mymodel~significantly surpasses OSEDiff across all evaluation criteria, highlighting the effectiveness of incorporating compression-aware visual embeddings into the diffusion process, as well as the elaborately designed training objectives. Additionally, \mymodel~outperforms MSD models, such as DiffBIR and SUPIR, in most metrics, despite requiring much fewer sampling steps. This suggests that \mymodel~can achieve high-quality restoration with greater efficiency.

Moreover, diffusion-based methods generally outperform CNN and Transformer-based approaches, particularly when handling highly compressed images (\eg, QF=5). This advantage primarily stems from the rich image generation priors of large-scale T2I diffusion models (\eg, StableDiffusion~\cite{rombach2022high}). These priors enable the models to compensate for the severe loss of visual information in highly compressed images, allowing for more faithful and perceptually realistic reconstructions.

\vspace{-0.1mm}
\noindent\textbf{Qualitative Results.} 
As shown in Fig.~\ref{fig:main_comp}, previous methods, including FBCNN~\cite{jiang2021towards}, JDEC~\cite{han2024jdec}, PromptCIR~\cite{li2024promptcir}, and OSEDiff~\cite{wu2024one}, can only partially mitigate JPEG artifact. However, noticeable compression artifact, such as blocky patterns, color banding, and grid-like distortions, still persist in their outputs. While multi-step DiffBIR is more effective in suppressing artifact, it tends to produce overly smooth results, sacrificing texture details in the process. In contrast, our \mymodel~effectively eliminates JPEG artifact while preserving fine-grained contents, such as fur patterns and architectural details. This highlights the model’s ability to recover complex details that are often lost due to heavy compression. Overall, our method strikes a balance between artifact removal and texture preservation, ensuring that the restored images retain both high perceptual quality and structural integrity. 

\vspace{-2mm}
\subsection{Complexity Analyses}
\vspace{-2mm}
Table~\ref{table:complexity} presents a comparison of model complexity, considering key factors such as the number of sampling steps (\#Step), parameter number (Params), multiply-accumulate operations (MACs), and inference time (Time). To ensure a fair assessment, all models are evaluated on a single NVIDIA A6000 GPU, except for SUPIR, which requires two A6000 GPUs due to its extensive model size. Notably, \mymodel~achieves a significant reduction in computational cost, outperforming MSD models by a substantial margin. This highlights its remarkable efficiency without compromising image quality. Moreover, \mymodel~requires fewer parameters compared to other diffusion based approaches, because we avoid complex auxiliary modules. Specifically, OSEDiff~\cite{wu2024one} employs DAPE~\cite{wu2024seesr} for textual prompt extraction. Meanwhile, DiffBIR~\cite{lin2024diffbir} and SUPIR~\cite{yu2024scaling} utilize ControlNet~\cite{zhang2023adding} to integrate low-quality (LQ) image information as the denosing condition. These auxiliary modules substantially increasing parameter counts. In contrast, our CaVE is a lightweight module that efficiently extracts compression priors with low computational overhead.

\begin{table}[t]
\footnotesize
\centering
\begin{center}
\vspace{.10mm}
\resizebox{1\columnwidth}{!}{
\begin{tabular}{l|c|c|c|c}
\toprule[0.15em]
\rowcolor{gray}
Method &  {\#Step} & Params (G) &  MACs (T) & Time (s)
\\
\midrule[0.15em]
DiffBIR~\cite{lin2024diffbir} & 50 & 1.52 & 188.24 & 50.81     
\\
SUPIR~\cite{yu2024scaling} & 50 & 4.49 & 464.29 & 24.33
\\
OSEDiff~\cite{wu2024one} & 1 & 1.40 & 10.39 & 0.65        
\\
\mymodel~(ours) & 1 & 1.00 & 9.46 & 0.57    
\\
\bottomrule[0.15em]
\end{tabular}}
\vspace{-3mm}
\caption{Complexity comparison among diffusion-based methods. Input image size is 1,024$\times$1,024 for inference.} 

\label{table:complexity}
\end{center}
\vspace{-6mm}
\end{table}

\subsection{Ablation Studies}
\begin{figure}[t]
\scriptsize
\centering
\begin{tabular}{cc}
\hspace{-0.56cm}

\begin{tabular}{c}
\includegraphics[width=0.14\textwidth, height=0.181\textwidth]{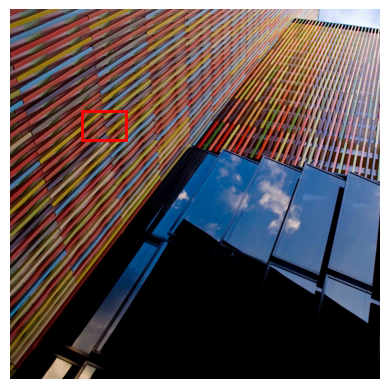}
\\
Urban100: img\_023 \\
\end{tabular}

\hspace{-5.6mm}

\begin{tabular}{ccc}
\includegraphics[width=0.114\textwidth]{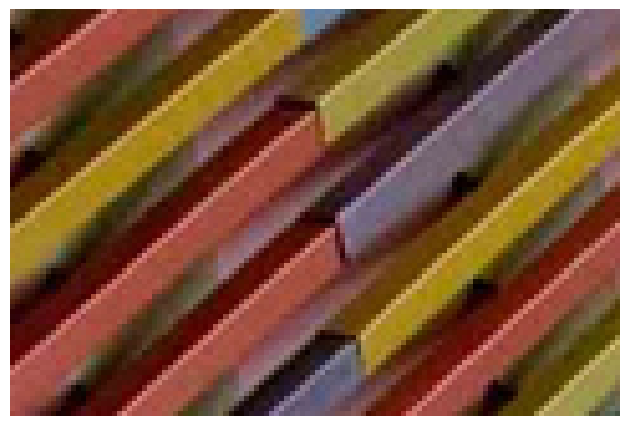} \hspace{-5mm} &
\includegraphics[width=0.114\textwidth]{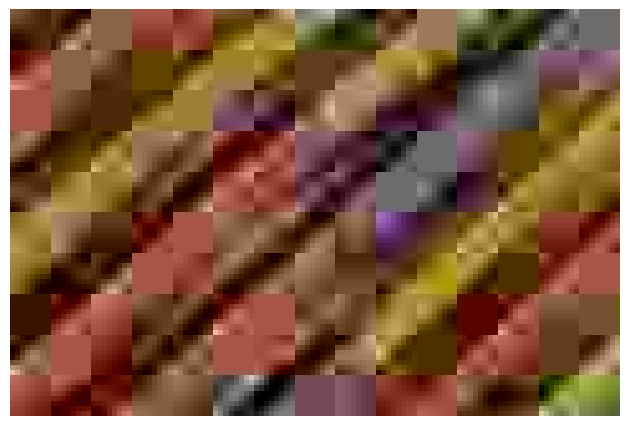} \hspace{-5mm} &
\includegraphics[width=0.114\textwidth]{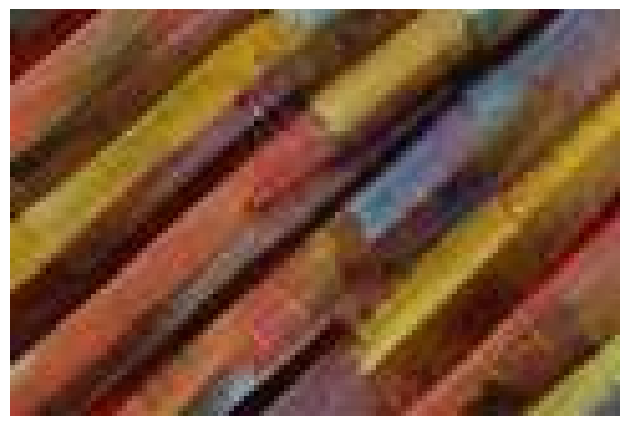} \hspace{-5mm}
\\
HQ \hspace{-5mm}
 &
JPEG (QF=5) \hspace{-5mm}  &
Empty String \hspace{-5mm}
\\

\includegraphics[width=0.114\textwidth]{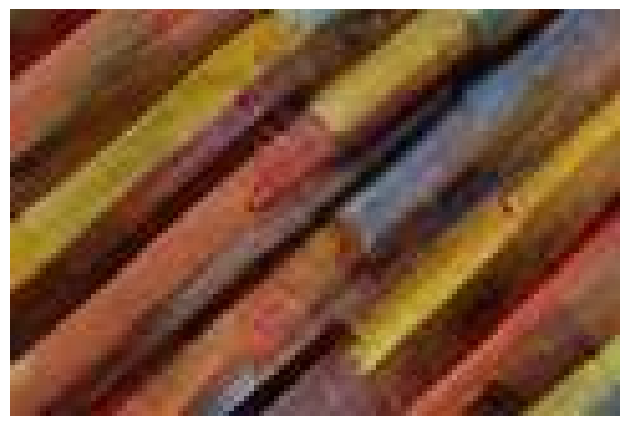} \hspace{-5mm} &
\includegraphics[width=0.114\textwidth]{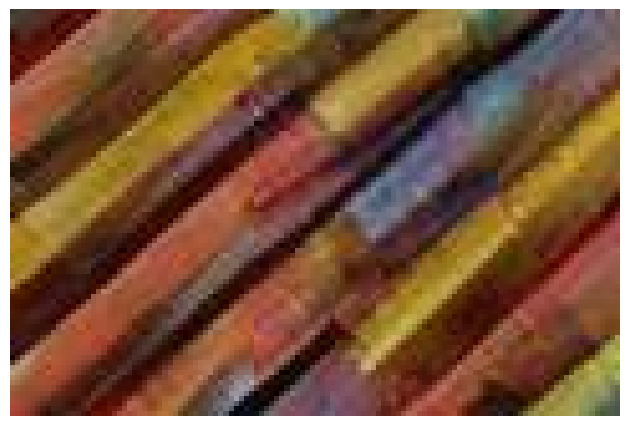} \hspace{-5mm} &
\includegraphics[width=0.114\textwidth]{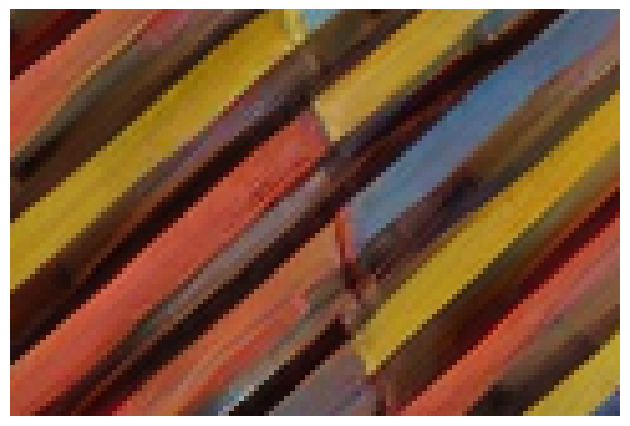} \hspace{-5mm}  
\\ 
Learnable \hspace{-5mm}  &
DAPE~\cite{wu2024seesr} \hspace{-5mm} &
CaVE~(ours) \hspace{-5mm}
\\

\end{tabular}

\end{tabular}
\vspace{-3mm}
\caption{Visual comparison of prompt embedding generation. We use Urban100 img\_023 as an example.}
\label{fig:embed}
\vspace{-5.5mm}

\end{figure}

\vspace{-0.5mm}
\noindent\textbf{Compression-aware Visual Embedder (CaVE).} We assess the effectiveness of different prompt methods, including empty strings, learnable embeddings, Degradation-Aware Prompt Extractor (DAPE)~\cite{wu2024seesr}, and CaVE. Table~\ref{table:embed} provides the results on LIVE-1 and DIV2K-Val test sets, focusing on challenging scenarios, where extreme compression leads to severe visual information loss. Our findings demonstrate that CaVE consistently outperforms other approaches across most evaluation metrics. In Fig.~\ref{fig:embed}, utilizing CaVE significantly reduces JPEG artifact, yielding noticeably more clear image reconstructions. Such visualization further shows the effect of CaVE.

\begin{table}[t]
\Large
\centering
\begin{center}
\vspace{.10mm}

\resizebox{1\columnwidth}{!}{
\begin{tabular}{l|ccc|ccc}
\toprule[0.15em]
\rowcolor{gray} & \multicolumn{3}{c|}{LIVE-1} & \multicolumn{3}{c}{DIV2K-Val} \\

\rowcolor{gray}
\multirow{-2}{*}{Method} & LPIPS$\downarrow$ & MUSIQ$\uparrow$ & M-IQA$\uparrow$ & LPIPS$\downarrow$ & MUSIQ$\uparrow$ & M-IQA$\uparrow$ \\
\midrule[0.15em]
Empty & 0.3485 & 62.56 & 0.3793 & 0.3241 & 56.68 & 0.3142 \\
Learnable & 0.3471 & \textcolor{iccvblue}{63.39} & \textcolor{iccvblue}{0.3900} & 0.3235 & 56.92 & \textcolor{iccvblue}{0.3262} \\
DAPE~\cite{wu2024seesr} & \textcolor{iccvblue}{0.3463} & 62.54 & 0.3793 & \textcolor{iccvblue}{0.3230} & \textcolor{iccvblue}{57.29} & 0.3240 \\
CaVE (ours) & \textcolor{red}{0.3426} & \textcolor{red}{67.13} & \textcolor{red}{0.4584} & \textcolor{red}{0.3179} & \textcolor{red}{61.83} & \textcolor{red}{0.3709} \\
\bottomrule[0.15em]
\end{tabular}}
\vspace{-3mm}
\caption{Ablation study of prompt embedding generation on LIVE-1 and DIV2K-val datasets. The best and second best results are colored with \textcolor{red}{red} and \textcolor{iccvblue}{blue}, respectively.} 
\label{table:embed}
\end{center}
\vspace{-7mm}
\end{table}

\noindent\textbf{Dual Learning.} To validate the effectiveness of our dual learning, we train CaVE with different learning strategies. Then we utilize CaVE to guide the reconstruction of \mymodel. As shown in Tab.~\ref{table:cavelearning}, the model trained with dual learning consistently achieves the best performance across all evaluation metrics. This suggests that dual learning paradigms enables CaVE to capture richer compression priors.

\begin{figure}[t]
    \begin{center}
    \begin{tabular}{cc}
    \hspace{-4mm}
        \includegraphics[width=0.5\columnwidth]{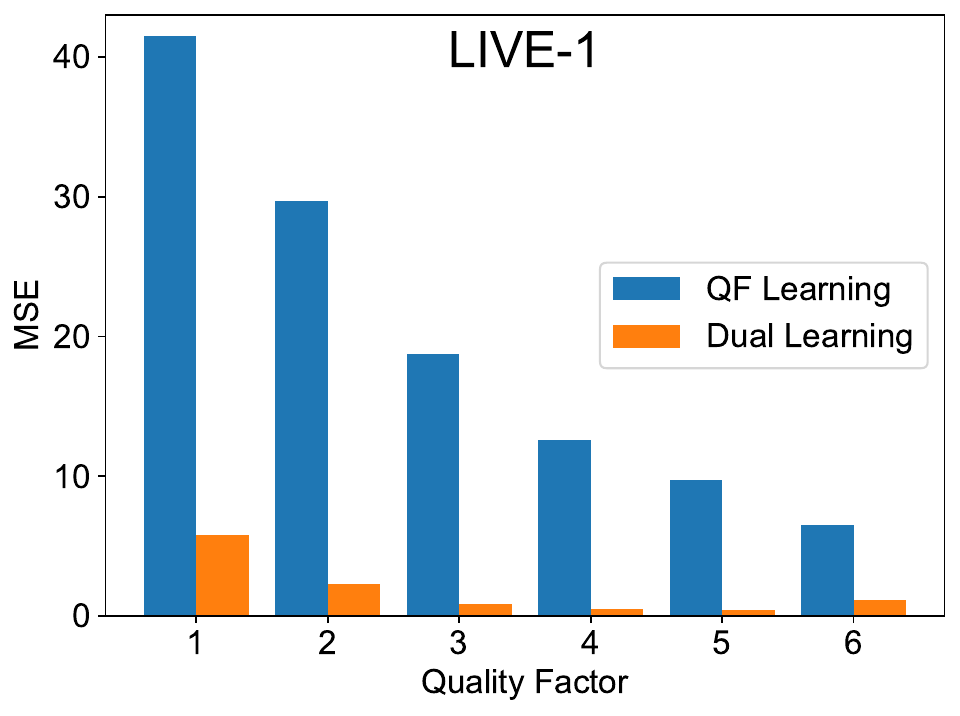}  &
        \hspace{-3.5mm}
        \includegraphics[width=0.5\columnwidth]{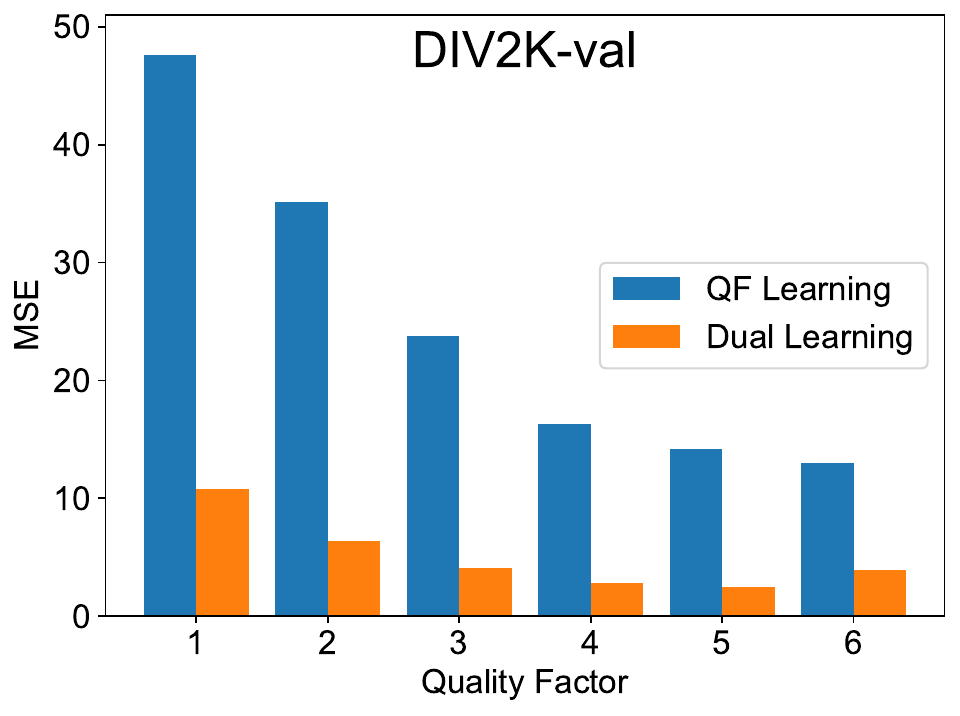} 
    \end{tabular}
    \end{center}
    \vspace{-8.5mm}
    \caption{Comparison of CaVE's QF prediction mean squared error (MSE) between explicit learning and dual learning on LIVE-1 and DIV2K-val datasets.}
    \label{fig:qf_pred}
    \vspace{-6mm} 
\end{figure}

Moreover, we evaluate CaVE's generalization ability by predicting QFs from low-quality images, specifically testing on QFs that are absent from the training set. As shown in Fig.~\ref{fig:qf_pred}, CaVE trained with explicit learning struggles to generalize to unseen QFs, whereas the dual learning approach significantly enhances its generalization capability.

\begin{table}[t]
\Large
\centering
\begin{center}
\vspace{.10mm}

\resizebox{1\columnwidth}{!}{
\begin{tabular}{l|ccc|ccc}
\toprule[0.15em]
\rowcolor{gray} & \multicolumn{3}{c|}{LIVE-1} & \multicolumn{3}{c}{DIV2K-Val} \\

\rowcolor{gray}
\multirow{-2}{*}{Type} & LPIPS$\downarrow$ & MUSIQ$\uparrow$ & M-IQA$\uparrow$ & LPIPS$\downarrow$ & MUSIQ$\uparrow$ & M-IQA$\uparrow$ \\
\midrule[0.15em]
Explicit & 0.3733 & 62.39 & 0.3828 & 0.3398 & 57.26 & 0.3202 \\
Implicit & \textcolor{iccvblue}{0.3436} & \textcolor{iccvblue}{64.74} & \textcolor{iccvblue}{0.4192} & \textcolor{iccvblue}{0.3208} & \textcolor{iccvblue}{58.74} & \textcolor{iccvblue}{0.3403} \\
Dual & \textcolor{red}{0.3426} & \textcolor{red}{67.13} & \textcolor{red}{0.4584} & \textcolor{red}{0.3179} & \textcolor{red}{61.83} & \textcolor{red}{0.3709} \\
\bottomrule[0.15em]
\end{tabular}}
\vspace{-3mm}
\caption{Ablation study of the dual learning strategy. The best and second best results are colored with \textcolor{red}{red} and \textcolor{iccvblue}{blue}, respectively.} 

\label{table:cavelearning}
\end{center}
\vspace{-7mm}
\end{table}

\noindent\textbf{\mymodel~Training Loss Functions.}
\mymodel~incorporates multiple loss functions during training, including MSE loss, DISTS loss, and GAN loss, to enhance overall model performance. To systematically evaluate the contribution of each component, we conduct extensive ablation studies. As shown in Tab.~\ref{table:ablation_loss}, training \mymodel~solely with $\mathcal{L}_{2}$ results in suboptimal perceptual metrics. However, incorporating $\mathcal{L}_{DISTS}$ significantly enhances \mymodel's performance. This highlights the pivotal role of DISTS loss in enhancing model's ability to achieve high-quality reconstructions. Furthermore, adding the GAN loss provides an additional performance boost, indicating its complementary effect in refining the reconstruction quality. 

\begin{table}[t]
    \centering
    \footnotesize
    \setlength{\tabcolsep}{0.5mm} 
    \newcolumntype{C}{>{\centering\arraybackslash}X}

    \newcolumntype{S}{>{\centering\arraybackslash}c}

    \begin{tabularx}{\columnwidth}{SSS|*{5}{C}} 
        \toprule[0.15em]
        \rowcolor{gray} $\mathcal{L}_2$ & $\mathcal{L}_{\text{DISTS}}$ & $\mathcal{L}_\mathcal{G}$ & LPIPS$\downarrow$ & DISTS$\downarrow$ & MUSIQ$\uparrow$ & M-IQA$\uparrow$& C-IQA$\uparrow$ \\
        \midrule[0.15em]
        \checkmark & &  & 0.3976 & 0.2509 & 59.79 & 0.3212 & 0.3085  \\
        \checkmark & \checkmark & & \textcolor{red}{0.2109} & \textcolor{iccvblue}{0.1229} & \textcolor{iccvblue}{72.69} & \textcolor{iccvblue}{0.5147} & \textcolor{iccvblue}{0.7119} \\ \checkmark & & \checkmark &  0.3899 & 0.2411 & 59.98 & 0.3301 & 0.3130 \\
        \checkmark & \checkmark & \checkmark & \textcolor{iccvblue}{0.2113} & \textcolor{red}{0.1147} & \textcolor{red}{73.34} & \textcolor{red}{0.5221} & \textcolor{red}{0.7317}  \\
        \bottomrule[0.15em]
    \end{tabularx}

    \vspace{-3mm}
    \caption{Ablation studies on different loss functions. The best and second best results are colored with \textcolor{red}{red} and \textcolor{iccvblue}{blue}, respectively.}
    \label{table:ablation_loss}
    \vspace{-4mm}
\end{table}

\vspace{-1mm}
\section{Conclusion}
\vspace{-1mm}
We propose \mymodel, a one-step diffusion model designed for efficient JPEG artifact removal. Our approach leverages the generative capabilities of text-to-image diffusion models to restore heavily JPEG-compressed images. By integrating a compression-aware visual embedder, we effectively utilize JPEG compression priors to guide the denoising process. This enables more accurate and visually coherent restorations. Beyond addressing the limitations of existing methods, \mymodel~establishes a novel pathway for using compression priors to guide diffusion models for JPEG artifact removal. Extensive experiments show the superiority of our \mymodel~over recent leading methods.

\section*{Acknowledgments}
This work was supported by Shanghai Municipal Science and Technology Major Project (2021SHZDZX0102) and the Fundamental Research Funds for the Central Universities.

{
    \small
    \bibliographystyle{ieeenat_fullname}
    \bibliography{main}

\begin{thebibliography}{56}
\providecommand{\natexlab}[1]{#1}
\providecommand{\url}[1]{\texttt{#1}}
\expandafter\ifx\csname urlstyle\endcsname\relax
  \providecommand{\doi}[1]{doi: #1}\else
  \providecommand{\doi}{doi: \begingroup \urlstyle{rm}\Url}\fi

\bibitem[Agustsson and Timofte(2017)]{agustsson2017ntire}
Eirikur Agustsson and Radu Timofte.
\newblock Ntire 2017 challenge on single image super-resolution: Dataset and study.
\newblock In \emph{CVPRW}, 2017.

\bibitem[Dhariwal and Nichol(2021)]{dhariwal2021diffusion}
Prafulla Dhariwal and Alexander Nichol.
\newblock Diffusion models beat gans on image synthesis.
\newblock \emph{NeurIPS}, 2021.

\bibitem[Ding et~al.(2020)Ding, Ma, Wang, and Simoncelli]{ding2020image}
Keyan Ding, Kede Ma, Shiqi Wang, and Eero~P Simoncelli.
\newblock Image quality assessment: Unifying structure and texture similarity.
\newblock \emph{IEEE TPAMI}, 2020.

\bibitem[Dong et~al.(2014)Dong, Loy, He, and Tang]{dong2014learning}
Chao Dong, Chen~Change Loy, Kaiming He, and Xiaoou Tang.
\newblock Learning a deep convolutional network for image super-resolution.
\newblock In \emph{ECCV}, 2014.

\bibitem[Dong et~al.(2015)Dong, Deng, Loy, and Tang]{dong2015compression}
Chao Dong, Yubin Deng, Chen~Change Loy, and Xiaoou Tang.
\newblock Compression artifacts reduction by a deep convolutional network.
\newblock In \emph{ICCV}, 2015.

\bibitem[Ehrlich et~al.(2020)Ehrlich, Davis, Lim, and Shrivastava]{ehrlich2020quantization}
Max Ehrlich, Larry Davis, Ser-Nam Lim, and Abhinav Shrivastava.
\newblock Quantization guided jpeg artifact correction.
\newblock In \emph{ECCV}, 2020.

\bibitem[Galteri et~al.(2017)Galteri, Seidenari, Bertini, and Del~Bimbo]{galteri2017deep}
Leonardo Galteri, Lorenzo Seidenari, Marco Bertini, and Alberto Del~Bimbo.
\newblock Deep generative adversarial compression artifact removal.
\newblock In \emph{ICCV}, 2017.

\bibitem[Galteri et~al.(2019)Galteri, Seidenari, Bertini, and Del~Bimbo]{galteri2019deep}
Leonardo Galteri, Lorenzo Seidenari, Marco Bertini, and Alberto Del~Bimbo.
\newblock Deep universal generative adversarial compression artifact removal.
\newblock \emph{IEEE TMM}, 2019.

\bibitem[Goodfellow et~al.(2020)Goodfellow, Pouget-Abadie, Mirza, Xu, Warde-Farley, Ozair, Courville, and Bengio]{goodfellow2020generative}
Ian Goodfellow, Jean Pouget-Abadie, Mehdi Mirza, Bing Xu, David Warde-Farley, Sherjil Ozair, Aaron Courville, and Yoshua Bengio.
\newblock Generative adversarial networks.
\newblock \emph{Communications of the ACM}, 2020.

\bibitem[Guo and Chao(2016)]{guo2016building}
Jun Guo and Hongyang Chao.
\newblock Building dual-domain representations for compression artifacts reduction.
\newblock In \emph{ECCV}, 2016.

\bibitem[Guo et~al.(2025)Guo, Ji, Chen, Liu, Liu, Rao, Li, Guo, and Zhang]{guo2025oscar}
Jinpei Guo, Yifei Ji, Zheng Chen, Kai Liu, Min Liu, Wang Rao, Wenbo Li, Yong Guo, and Yulun Zhang.
\newblock Oscar: One-step diffusion codec across multiple bit-rates.
\newblock \emph{arXiv preprint arXiv:2505.16091}, 2025.

\bibitem[Han et~al.(2024)Han, Im, Kim, and Jin]{han2024jdec}
Woo~Kyoung Han, Sunghoon Im, Jaedeok Kim, and Kyong~Hwan Jin.
\newblock Jdec: Jpeg decoding via enhanced continuous cosine coefficients.
\newblock In \emph{CVPR}, 2024.

\bibitem[Ho et~al.(2020)Ho, Jain, and Abbeel]{ho2020denoising}
Jonathan Ho, Ajay Jain, and Pieter Abbeel.
\newblock Denoising diffusion probabilistic models.
\newblock \emph{NeurIPS}, 2020.

\bibitem[Hu et~al.(2021)Hu, Shen, Wallis, Allen-Zhu, Li, Wang, Wang, and Chen]{hu2021lora}
Edward~J Hu, Yelong Shen, Phillip Wallis, Zeyuan Allen-Zhu, Yuanzhi Li, Shean Wang, Lu Wang, and Weizhu Chen.
\newblock Lora: Low-rank adaptation of large language models.
\newblock \emph{arXiv preprint arXiv:2106.09685}, 2021.

\bibitem[Huang et~al.(2015)Huang, Singh, and Ahuja]{huang2015single}
Jia-Bin Huang, Abhishek Singh, and Narendra Ahuja.
\newblock Single image super-resolution from transformed self-exemplars.
\newblock In \emph{CVPR}, 2015.

\bibitem[Jiang et~al.(2021)Jiang, Zhang, and Timofte]{jiang2021towards}
Jiaxi Jiang, Kai Zhang, and Radu Timofte.
\newblock Towards flexible blind jpeg artifacts removal.
\newblock In \emph{ICCV}, 2021.

\bibitem[Jiang et~al.(2024)Jiang, Zhang, Xue, and Gu]{jiang2024autodir}
Yitong Jiang, Zhaoyang Zhang, Tianfan Xue, and Jinwei Gu.
\newblock Autodir: Automatic all-in-one image restoration with latent diffusion.
\newblock In \emph{ECCV}, 2024.

\bibitem[Ke et~al.(2021)Ke, Wang, Wang, Milanfar, and Yang]{ke2021musiq}
Junjie Ke, Qifei Wang, Yilin Wang, Peyman Milanfar, and Feng Yang.
\newblock Musiq: Multi-scale image quality transformer.
\newblock In \emph{ICCV}, 2021.

\bibitem[Kim et~al.(2020)Kim, Soh, and Cho]{kim2020agarnet}
Yoonsik Kim, Jae~Woong Soh, and Nam~Ik Cho.
\newblock Agarnet: Adaptively gated jpeg compression artifacts removal network for a wide range quality factor.
\newblock \emph{IEEE Access}, 2020.

\bibitem[Kingma(2014)]{kingma2014adam}
Diederik~P Kingma.
\newblock Adam: A method for stochastic optimization.
\newblock \emph{arXiv preprint arXiv:1412.6980}, 2014.

\bibitem[Li et~al.(2024{\natexlab{a}})Li, Li, Lu, Feng, Guo, Zhao, Zhang, and Chen]{li2024promptcir}
Bingchen Li, Xin Li, Yiting Lu, Ruoyu Feng, Mengxi Guo, Shijie Zhao, Li Zhang, and Zhibo Chen.
\newblock Promptcir: Blind compressed image restoration with prompt learning.
\newblock \emph{arXiv preprint arXiv:2404.17433}, 2024{\natexlab{a}}.

\bibitem[Li et~al.(2020)Li, Wang, Xie, and Ma]{li2020learning}
Jianwei Li, Yongtao Wang, Haihua Xie, and Kai-Kuang Ma.
\newblock Learning a single model with a wide range of quality factors for jpeg image artifacts removal.
\newblock \emph{IEEE TIP}, 2020.

\bibitem[Li et~al.(2024{\natexlab{b}})Li, Cao, Zou, Su, Yuan, Zhang, Guo, and Yang]{li2024distillation}
Jianze Li, Jiezhang Cao, Zichen Zou, Xiongfei Su, Xin Yuan, Yulun Zhang, Yong Guo, and Xiaokang Yang.
\newblock Distillation-free one-step diffusion for real-world image super-resolution.
\newblock \emph{arXiv preprint arXiv:2410.04224}, 2024{\natexlab{b}}.

\bibitem[Li et~al.(2023)Li, Zhang, Liang, Cao, Liu, Gong, Zhang, Tang, Liu, Demandolx, et~al.]{li2023lsdir}
Yawei Li, Kai Zhang, Jingyun Liang, Jiezhang Cao, Ce Liu, Rui Gong, Yulun Zhang, Hao Tang, Yun Liu, Denis Demandolx, et~al.
\newblock Lsdir: A large scale dataset for image restoration.
\newblock In \emph{CVPR}, 2023.

\bibitem[Liang et~al.(2021)Liang, Cao, Sun, Zhang, Van~Gool, and Timofte]{liang2021swinir}
Jingyun Liang, Jiezhang Cao, Guolei Sun, Kai Zhang, Luc Van~Gool, and Radu Timofte.
\newblock Swinir: Image restoration using swin transformer.
\newblock In \emph{ICCV}, 2021.

\bibitem[Lin et~al.(2024)Lin, He, Chen, Lyu, Dai, Yu, Qiao, Ouyang, and Dong]{lin2024diffbir}
Xinqi Lin, Jingwen He, Ziyan Chen, Zhaoyang Lyu, Bo Dai, Fanghua Yu, Yu Qiao, Wanli Ouyang, and Chao Dong.
\newblock Diffbir: Toward blind image restoration with generative diffusion prior.
\newblock In \emph{ECCV}, 2024.

\bibitem[Loshchilov(2017)]{loshchilov2017decoupled}
I Loshchilov.
\newblock Decoupled weight decay regularization.
\newblock \emph{arXiv preprint arXiv:1711.05101}, 2017.

\bibitem[Paszke et~al.(2019)Paszke, Gross, Massa, Lerer, Bradbury, Chanan, Killeen, Lin, Gimelshein, Antiga, et~al.]{paszke2019pytorch}
Adam Paszke, Sam Gross, Francisco Massa, Adam Lerer, James Bradbury, Gregory Chanan, Trevor Killeen, Zeming Lin, Natalia Gimelshein, Luca Antiga, et~al.
\newblock Pytorch: An imperative style, high-performance deep learning library.
\newblock In \emph{NeurIPS}, 2019.

\bibitem[Podell et~al.(2023)Podell, English, Lacey, Blattmann, Dockhorn, M{\"u}ller, Penna, and Rombach]{podell2023sdxl}
Dustin Podell, Zion English, Kyle Lacey, Andreas Blattmann, Tim Dockhorn, Jonas M{\"u}ller, Joe Penna, and Robin Rombach.
\newblock Sdxl: Improving latent diffusion models for high-resolution image synthesis.
\newblock \emph{arXiv preprint arXiv:2307.01952}, 2023.

\bibitem[Rombach et~al.(2022)Rombach, Blattmann, Lorenz, Esser, and Ommer]{rombach2022high}
Robin Rombach, Andreas Blattmann, Dominik Lorenz, Patrick Esser, and Bj{\"o}rn Ommer.
\newblock High-resolution image synthesis with latent diffusion models.
\newblock In \emph{CVPR}, 2022.

\bibitem[Ronneberger et~al.(2015)Ronneberger, Fischer, and Brox]{ronneberger2015u}
Olaf Ronneberger, Philipp Fischer, and Thomas Brox.
\newblock U-net: Convolutional networks for biomedical image segmentation.
\newblock In \emph{MICCAI}, 2015.

\bibitem[Saharia et~al.(2022)Saharia, Chan, Saxena, Li, Whang, Denton, Ghasemipour, Gontijo~Lopes, Karagol~Ayan, Salimans, et~al.]{saharia2022photorealistic}
Chitwan Saharia, William Chan, Saurabh Saxena, Lala Li, Jay Whang, Emily~L Denton, Kamyar Ghasemipour, Raphael Gontijo~Lopes, Burcu Karagol~Ayan, Tim Salimans, et~al.
\newblock Photorealistic text-to-image diffusion models with deep language understanding.
\newblock \emph{NeurIPS}, 2022.

\bibitem[Sheikh(2005)]{sheikh2005live}
H Sheikh.
\newblock Live image quality assessment database release 2.
\newblock \emph{http://live. ece. utexas. edu/research/quality}, 2005.

\bibitem[Song et~al.(2020{\natexlab{a}})Song, Meng, and Ermon]{song2020denoising}
Jiaming Song, Chenlin Meng, and Stefano Ermon.
\newblock Denoising diffusion implicit models.
\newblock \emph{arXiv preprint arXiv:2010.02502}, 2020{\natexlab{a}}.

\bibitem[Song et~al.(2020{\natexlab{b}})Song, Sohl-Dickstein, Kingma, Kumar, Ermon, and Poole]{song2020score}
Yang Song, Jascha Sohl-Dickstein, Diederik~P Kingma, Abhishek Kumar, Stefano Ermon, and Ben Poole.
\newblock Score-based generative modeling through stochastic differential equations.
\newblock \emph{arXiv preprint arXiv:2011.13456}, 2020{\natexlab{b}}.

\bibitem[Timofte et~al.(2017)Timofte, Agustsson, Van~Gool, Yang, and Zhang]{timofte2017ntire}
Radu Timofte, Eirikur Agustsson, Luc Van~Gool, Ming-Hsuan Yang, and Lei Zhang.
\newblock Ntire 2017 challenge on single image super-resolution: Methods and results.
\newblock In \emph{CVPRW}, 2017.

\bibitem[Van~der Maaten and Hinton(2008)]{van2008visualizing}
Laurens Van~der Maaten and Geoffrey Hinton.
\newblock Visualizing data using t-sne.
\newblock \emph{JMLR}, 2008.

\bibitem[Wallace(1991)]{wallace1991jpeg}
Gregory~K Wallace.
\newblock The jpeg still picture compression standard.
\newblock \emph{Communications of the ACM}, 1991.

\bibitem[Wang et~al.(2023)Wang, Chan, and Loy]{wang2023exploring}
Jianyi Wang, Kelvin~CK Chan, and Chen~Change Loy.
\newblock Exploring clip for assessing the look and feel of images.
\newblock In \emph{AAAI}, 2023.

\bibitem[Wang et~al.(2024{\natexlab{a}})Wang, Yue, Zhou, Chan, and Loy]{wang2024exploiting}
Jianyi Wang, Zongsheng Yue, Shangchen Zhou, Kelvin~CK Chan, and Chen~Change Loy.
\newblock Exploiting diffusion prior for real-world image super-resolution.
\newblock \emph{IJCV}, 2024{\natexlab{a}}.

\bibitem[Wang et~al.(2021)Wang, Fu, Sun, and Zha]{wang2021jpeg}
Menglu Wang, Xueyang Fu, Zepei Sun, and Zheng-Jun Zha.
\newblock Jpeg artifacts removal via compression quality ranker-guided networks.
\newblock In \emph{IJCAI}, 2021.

\bibitem[Wang et~al.(2022)Wang, Fu, Zhu, and Zha]{wang2022jpeg}
Xi Wang, Xueyang Fu, Yurui Zhu, and Zheng-Jun Zha.
\newblock Jpeg artifacts removal via contrastive representation learning.
\newblock In \emph{ECCV}, 2022.

\bibitem[Wang et~al.(2024{\natexlab{b}})Wang, Yang, Chen, Wang, Guo, Chau, Liu, Qiao, Kot, and Wen]{wang2024sinsr}
Yufei Wang, Wenhan Yang, Xinyuan Chen, Yaohui Wang, Lanqing Guo, Lap-Pui Chau, Ziwei Liu, Yu Qiao, Alex~C Kot, and Bihan Wen.
\newblock Sinsr: diffusion-based image super-resolution in a single step.
\newblock In \emph{CVPR}, 2024{\natexlab{b}}.

\bibitem[Wu et~al.(2024{\natexlab{a}})Wu, Sun, Ma, and Zhang]{wu2024one}
Rongyuan Wu, Lingchen Sun, Zhiyuan Ma, and Lei Zhang.
\newblock One-step effective diffusion network for real-world image super-resolution.
\newblock \emph{arXiv preprint arXiv:2406.08177}, 2024{\natexlab{a}}.

\bibitem[Wu et~al.(2024{\natexlab{b}})Wu, Yang, Sun, Zhang, Li, and Zhang]{wu2024seesr}
Rongyuan Wu, Tao Yang, Lingchen Sun, Zhengqiang Zhang, Shuai Li, and Lei Zhang.
\newblock Seesr: Towards semantics-aware real-world image super-resolution.
\newblock In \emph{CVPR}, 2024{\natexlab{b}}.

\bibitem[Yang et~al.(2022)Yang, Wu, Shi, Lao, Gong, Cao, Wang, and Yang]{yang2022maniqa}
Sidi Yang, Tianhe Wu, Shuwei Shi, Shanshan Lao, Yuan Gong, Mingdeng Cao, Jiahao Wang, and Yujiu Yang.
\newblock Maniqa: Multi-dimension attention network for no-reference image quality assessment.
\newblock In \emph{CVPR}, 2022.

\bibitem[Yang et~al.(2024)Yang, Wu, Ren, Xie, and Zhang]{yang2024pixel}
Tao Yang, Rongyuan Wu, Peiran Ren, Xuansong Xie, and Lei Zhang.
\newblock Pixel-aware stable diffusion for realistic image super-resolution and personalized stylization.
\newblock In \emph{ECCV}, 2024.

\bibitem[Yin et~al.(2024{\natexlab{a}})Yin, Gharbi, Park, Zhang, Shechtman, Durand, and Freeman]{yin2024dmd2}
Tianwei Yin, Micha{\"e}l Gharbi, Taesung Park, Richard Zhang, Eli Shechtman, Fredo Durand, and William~T Freeman.
\newblock Improved distribution matching distillation for fast image synthesis.
\newblock \emph{arXiv preprint arXiv:2405.14867}, 2024{\natexlab{a}}.

\bibitem[Yin et~al.(2024{\natexlab{b}})Yin, Gharbi, Zhang, Shechtman, Durand, Freeman, and Park]{yin2024dmd}
Tianwei Yin, Micha{\"e}l Gharbi, Richard Zhang, Eli Shechtman, Fredo Durand, William~T Freeman, and Taesung Park.
\newblock One-step diffusion with distribution matching distillation.
\newblock In \emph{CVPR}, 2024{\natexlab{b}}.

\bibitem[Yu et~al.(2024)Yu, Gu, Li, Hu, Kong, Wang, He, Qiao, and Dong]{yu2024scaling}
Fanghua Yu, Jinjin Gu, Zheyuan Li, Jinfan Hu, Xiangtao Kong, Xintao Wang, Jingwen He, Yu Qiao, and Chao Dong.
\newblock Scaling up to excellence: Practicing model scaling for photo-realistic image restoration in the wild.
\newblock In \emph{CVPR}, 2024.

\bibitem[Yue et~al.(2024)Yue, Wang, and Loy]{yue2024resshift}
Zongsheng Yue, Jianyi Wang, and Chen~Change Loy.
\newblock Resshift: Efficient diffusion model for image super-resolution by residual shifting.
\newblock \emph{NeurIPS}, 2024.

\bibitem[Zhang et~al.(2023)Zhang, Rao, and Agrawala]{zhang2023adding}
Lvmin Zhang, Anyi Rao, and Maneesh Agrawala.
\newblock Adding conditional control to text-to-image diffusion models.
\newblock In \emph{ICCV}, 2023.

\bibitem[Zhang et~al.(2018{\natexlab{a}})Zhang, Isola, Efros, Shechtman, and Wang]{zhang2018unreasonable}
Richard Zhang, Phillip Isola, Alexei~A Efros, Eli Shechtman, and Oliver Wang.
\newblock The unreasonable effectiveness of deep features as a perceptual metric.
\newblock In \emph{CVPR}, 2018{\natexlab{a}}.

\bibitem[Zhang et~al.(2018{\natexlab{b}})Zhang, Yang, Hu, and Liu]{zhang2018dmcnn}
Xiaoshuai Zhang, Wenhan Yang, Yueyu Hu, and Jiaying Liu.
\newblock Dmcnn: Dual-domain multi-scale convolutional neural network for compression artifacts removal.
\newblock In \emph{ICIP}, 2018{\natexlab{b}}.

\bibitem[Zhang et~al.(2019)Zhang, Li, Li, Zhong, and Fu]{zhang2019residual}
Yulun Zhang, Kunpeng Li, Kai Li, Bineng Zhong, and Yun Fu.
\newblock Residual non-local attention networks for image restoration.
\newblock \emph{arXiv preprint arXiv:1903.10082}, 2019.

\bibitem[Zheng et~al.(2019)Zheng, Chen, Tian, Zhou, and Liu]{zheng2019implicit}
Bolun Zheng, Yaowu Chen, Xiang Tian, Fan Zhou, and Xuesong Liu.
\newblock Implicit dual-domain convolutional network for robust color image compression artifact reduction.
\newblock \emph{IEEE TCSVT}, 2019.

\end{thebibliography}
}

\end{document}